\documentclass{article}

\PassOptionsToPackage{numbers, compress}{natbib}

\usepackage[preprint]{neurips_2023}

\usepackage[utf8]{inputenc} 
\usepackage[T1]{fontenc}    
\usepackage{hyperref}       
\usepackage{url}            
\usepackage{booktabs}       
\usepackage{amsfonts}       
\usepackage{nicefrac}       
\usepackage{microtype}      
\usepackage{xcolor}         

\usepackage{graphicx}
\usepackage{tabularx}
\usepackage{subfigure}
\usepackage{inconsolata}            
\usepackage{lipsum}                 
\usepackage{tikz}
\usepackage[many]{tcolorbox}
\usepackage{textcomp}

\usepackage{microtype}              
\usepackage{nicefrac}               
\usepackage{todonotes}              
\usepackage{titlesec}               
\usepackage{wrapfig}                
\usepackage{multirow}               
\usepackage{floatrow}
\usepackage{amsmath}
\usepackage{amssymb}
\usepackage{mathtools}
\usepackage{amsthm}
\usepackage{pdfx}
\usepackage{adjustbox}
\usepackage{scrextend}

\usepackage[linesnumbered,lined,boxed,commentsnumbered,ruled]{algorithm2e}
\definecolor{backcolour}{rgb}{0.95,0.95,0.95}
\definecolor{coco}{rgb}{0.6, 0.69, 0.98}

\tcbset{prompt/.style={
    enhanced,
    size=fbox,
    boxrule=3pt,
    arc=2mm,
    auto outer arc,
    left=10pt,
    right=10pt,
    top=10pt,
    bottom=10pt,
    fontupper=\fontfamily{cmtt}\selectfont, 
    colback=CB_gray!10,
    colframe=CB_gray!25,
    coltitle=CB_gray!100, 
}}
\definecolor{CB_gray}{gray}{0.5}

\def\Snospace~{\S{}}

\hypersetup{
    colorlinks=true,
    linkcolor=orange,
    filecolor=magenta,      
    urlcolor=orange,
    citecolor=orange,
}

\title{ Strategic Reasoning with Language Models}

\author{%
  Kanishk Gandhi \quad Dorsa Sadigh \quad Noah D. Goodman\\
  Stanford University \\
  \texttt{kanishk.gandhi@stanford.edu}
}

\begin{document}

\maketitle

\begin{abstract}
\vspace{-2mm}
Strategic reasoning enables agents to cooperate, communicate, and compete with other agents in diverse situations.
Existing approaches to solving strategic games rely on extensive training, yielding strategies that do not generalize to new scenarios or games without retraining. Large Language Models (LLMs), with their ability to comprehend and generate complex, context-rich language, could prove powerful as tools for strategic gameplay. This paper introduces an approach that uses pretrained LLMs with few-shot chain-of-thought examples to enable strategic reasoning for AI agents. Our approach uses systematically generated demonstrations of reasoning about states, values, and beliefs to prompt the model. Using extensive variations of simple matrix games, we show that strategies that are derived based on systematically generated prompts generalize almost perfectly to new game structures, alternate objectives, and hidden information. Additionally, we demonstrate our approach can lead to human-like negotiation strategies in realistic scenarios without any extra training or fine-tuning. Our results highlight the ability of LLMs, guided by systematic reasoning demonstrations, to adapt and excel in diverse strategic scenarios.\footnote{\label{fn:web}Project Website: \href{https://sites.google.com/view/strategic-reasoning-llms/}{https://sites.google.com/view/strategic-reasoning-llms/}}


\end{abstract}
\vspace{-4mm}
\section{Introduction}
\vspace{-3mm}
\label{sec:introduction}

Advances in game-playing AI, for games such as Chess, Go \citep{silver2016mastering,silver2018general}, Poker \citep{moravvcik2017deepstack, brown2018superhuman}, and more recently diplomacy \citep{bakhtin2021no,bakhtin2022human, kramar2022negotiation}, have shown how algorithms trained with a combination of imitation, reinforcement learning, and planning can be used for strategic reasoning. Despite the increasing sophistication of these models, they are limited in their ability to generalize to new strategic scenarios because they are often extensively and solely trained on a given target game. Humans, on the other hand, can readily adapt to new scenarios: changes to the rules of a game (\textit{in chess, where the rules are changed so that the knight moves in a line});
different styles of play (an expert player can be asked to play aggressively---\textit{like a sacrificial player such as Mikhail Tal}---or with a  low-risk approach---\textit{a positional player like Karpov}); alternate goals (such as achieving the lowest score or even with some arbitrary definition of score---\textit{the first person to give a check wins the game})\citep{lake2017building}.  

This paper aims to address the limitation of existing AI algorithms in their ability \emph{to adapt to new contexts} by exploring the potential of language models to engage in strategic reasoning—the ability to foresee potential actions of others in the pursuit of possibly conflicting objectives, and to devise optimal strategies accordingly. This concept, central to game theory, entails reasoning about the interplay between multiple agents with divergent interests. 
Large language models (LLMs) have recently been shown to express human-like strategies \citep{aher2022using,kwon2023reward} and flexibility in reasoning, potentially understanding nuanced and contextual information \citep{suzgun2022challenging}. Further, since language models are trained on a variety of data sources, they can be adapted to different tasks and environments, making them suitable for flexible reasoning and potentially generalization to new scenarios. Despite these successes, LLMs can however be brittle and unreliable in their reasoning, especially when reasoning about agents, social contexts \citep{sap2022neural} and planning \citep{valmeekam2022large}. 

To enable LLMs to reason strategically, flexibly and reliably, we propose an approach that shows models how to search through states, evaluate actions and their effects, and form beliefs by using systematically generated demonstrations of strategic reasoning. The LLM can then generalize to new scenarios through few-shot in-context examples of these systematically generated prompts.
To capture human-like strategic reasoning, an agent needs to 1) search through the space of states and actions: for example, \textit{a bot that negotiates with a vendor must understand the space of inventory and how its offers will affect the negotiation}, 2) assign values to these states and actions: \textit{the bot must understand which items are valuable to it, and what the vendor values}, 3) form beliefs about the partially-observable world: \textit{based on the vendor's actions, the bot must infer how much the vendor values the items}. 
We develop an automated ``prompt compiler'' that can be used to systematically generate these demonstrations. The demonstrations generated by our compiler structure the chain-of-thought reasoning prompts~\citep{nye2022show,wei2022chain} that  bias the language model towards more flexible reasoning before selecting an action.

In the remainder of the paper, we first present our approach for a prompt compiler that systematically generates strategic reasoning prompts. We then evaluate it through experiments with matrix games and negotiation games. We use scenarios with new game structures, alternate objectives, partial information, and communication to demonstrate the model's ability to flexibly search, assign values and form beliefs in new contexts. Finally, we apply our approach to a realistic negotiation task \citep{lewis2017deal} and create a human-like negotiation agent without any prior training. Our results demonstrate the potential of language models for flexible strategic reasoning and their ability to generalize to new scenarios with few or no additional examples.

\vspace{-2mm}

\section{Related Work}
\vspace{-2mm}

\label{sec:related-work}
We review literature in two complementary areas: the progress of strategic reasoning with AI agents, particularly in game-playing environments, and the advancement of reasoning capabilities in LLMs.

\textbf{Strategic Reasoning with AI Agents.} Progress in game-playing AI, driven by breakthroughs in reinforcement learning (RL), self-play, and integrating them with tree search have led to successful strategic agents for Chess, Go, Starcraft, Poker and DOTA \citep{silver2016mastering,silver2018general,vinyals2019grandmaster,moravvcik2017deepstack, berner2019dota}. These approaches showed how self-play and reinforcement learning (RL) could help create policies that outperform humans. However, they were limited in producing agents that were adept at adapting to novel situations, such as new rules or objectives \citep{lake2017building}. Recently, Cicero \citep{bakhtin2022human} demonstrated how language models could be used to create versatile agents capable of interacting and negotiating with humans through dialogue, by combining techniques in strategic reasoning and language modelling to create a dialogue-based agent that could play Diplomacy. Despite their success, Cicero employed a separate strategic planner, using an LLM only for translating predictions to dialogue (and some implicit planning). This limits the agent's flexibility; it would have to be retrained to be successful in a new version of the game or if a rule in the game changed. In this paper, we explore how we can guide the language model through systematic prompts so that it can be used as a flexible strategic planner.

\textbf{Reasoning with Language Models.} Large language models have been shown to be successful at reasoning \citep{suzgun2022challenging} in a variety of contexts, especially when paired with prompting techniques that allow them to think through their steps \citep{nye2022show,wei2022chain}. Other techniques have shown how reasoning can further be improved by breaking down the steps in a problem \citep{kojima2022large}, combining language models as modules or cascades \citep{dohan2022language, primer2022}, by fine-tuning/ bootstrapping \citep{zelikman2022star, alpaca} its own reasoning and by tuning through human feedback \citep{ouyang2022training}. In spite of these successes, LLMs are limited in their reasoning about agents \citep{sap2022neural} and there have been few attempts to apply these models to complex strategic reasoning tasks. In this paper, we draw on these advances to amplify the reasoning capabilities of  LLMs and design a method that systematically prompts them to be flexible, reliable strategic agents.

\begin{figure*}[btp]
    \centering
    \includegraphics[width=0.98\textwidth]{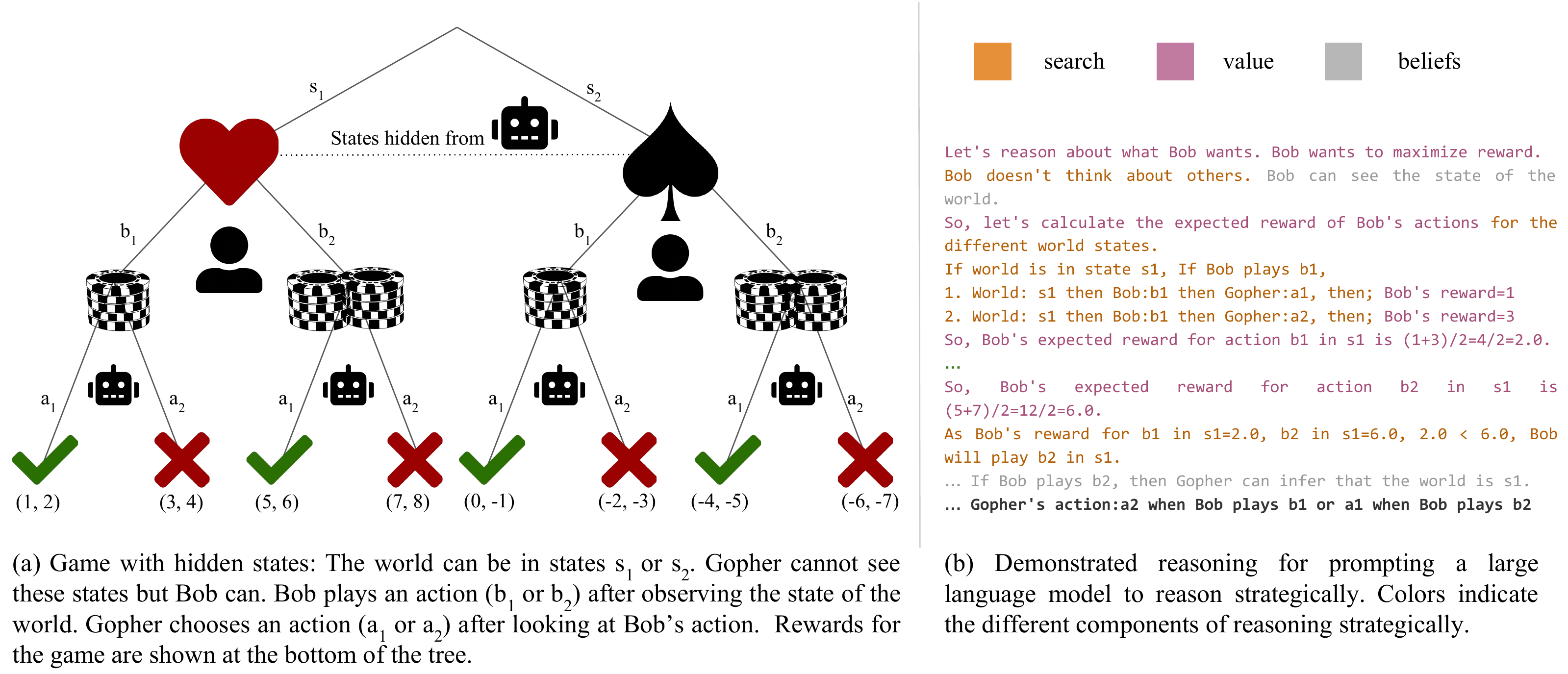}
    \caption{A partially observable game with two players Gopher and Bob is shown in (a), and the corresponding demonstration for strategic reasoning is shown in (b).}
    \vspace{-2mm}
    \label{fig:front}
\end{figure*}
\vspace{-3mm}
\section{Strategic Reasoning with Language Models}
\vspace{-2mm}
\label{sec:methods}

The key idea of our method is to guide the decision making of language models towards better strategic reasoning and enable them to generalize reliably, and flexibly to new domains.
We provide a systematic approach for generating prompts that incorporate structure based on common strategic reasoning techniques. 
Specifically, we will consider prompting strategies based on \emph{search}, \emph{value assignment}, and \emph{belief tracking}. To demonstrate our approach and prompting strategies, let us first describe two game settings that we study in this work: matrix games and negotiation games.
These games can be represented in the form of a multi-agent MDP. For a game with N agents, $S$ represents the set of world states, $A = \{A_i , \: \forall i \in N \}$ represents the set of actions taken by each agent, $R_i: S\times A_1\times\dots \times A_N \rightarrow \mathbb{R}$ is the set of rewards, and $P: S \times A_1\times\dots \times A_N \rightarrow S$ is the transition probability matrix. In addition, in a partially observable setting, parts of the state space or the rewards are not accessible to some or all of the agents.
 
\noindent \underline {\emph{Matrix Games:}} Game theorists have devised a vast array of simple \emph{matrix games} to exemplify rational strategic behavior; these games involve players acting to maximize self-interest while possessing perfect foresight of others' strategies. Insights from these games have found applications in diverse fields such as economics, politics, and evolutionary biology. In matrix games, the state space, action space and rewards are defined by the size and values of the payoff matrix; a $2\times 2$ matrix game has 2 players, where each player can take two actions. We vary the complexity of matrix games by changing the rewards, the objectives of the players, the action space, the number of players, the observability of the game and by adding multiple turns to a game.

\noindent \underline{\emph{Negotiation Games:}} In the negotiation setting, we use the Deal or No Deal game \citep{lewis2017deal}. Each episode in this game has two players that must split a pot of three types of items (hats, books and balls) with each other. The players value these items differently and do not know the other player's values. Across episodes, the number of each of the items and the players' values for the items is varied. 
 
\vspace{-3mm}

\subsection{Incorporating Structure in Strategic Reasoning Prompts}
\vspace{-2mm}

To enable a large language model to solve strategic reasoning games and generalize to its new instances, we propose an in-context learning approach, where demonstrations of solving strategic games are provided in the prompt. Our insight is that this prompt needs to incorporate \emph{structure} for reasoning about the strategic games. enabling LLMs to reliably leverage cognitive tools useful for strategic reasoning such as search, value assignment, and belief modeling.

Our prompt compiler takes an example game and automatically generates an example of strategic reasoning. The compiler decomposes this example into components for search, value assignment, and belief tracking. In matrix games, the examples are generated for simple and restricted games allowing for systematically testing for generalization capabilities to novel scenarios.
On the other hand, in the negotiation domain, the demonstrations are provided by a human player. These human demonstrations are then annotated with reasoning steps corresponding to search, value assignment and belief tracking by an expert demonstrator. In both settings, the compiler and the annotations leverage the decomposed games as contextual prompts to steer the behavior of the language model in a new strategic game. In \autoref{fig:front}, when the state of the world, hearts or spades is hidden from the player, we show how the reasoning is decomposed into search, value and belief formation to find the best response.

\textbf{Prompting Strategy: Search.} Searching flexibly is necessary for generalizing to changes in parts of a game, such as its rules, the number of players, or the action space. 
When possible, we perform an exhaustive search  by exploring the values of all leaf nodes \footnote{ The terminal points of a game tree that have no subsequent decisions/actions, associated with payoffs.
} in the game tree \footnote{A standard tree structure that represents the game states/scenarios and possible transitions between them based on players' actions/decisions and the rules of the game.
} (\autoref{alg:exhaustive}:3,9). 
For example, in a matrix game a player must iterate through the actions of the other players to find the right states to evaluate, ``\textit{If player 1 plays $a_1$ and player 2 plays $a_2$ then ...}''. 
We automatically generate this prompt (as shown in \autoref{fig:front}~b) to demonstrate search (see example prompt: \autoref{prompt:mat}).

\begin{figure}[t]
    \centering
    \begin{adjustbox}{max width=\textwidth}
    \begin{tabular}{@{}c@{\hspace{0.05\textwidth}}c@{}}
        \begin{minipage}[t]{0.55\textwidth}
\begin{algorithm}[H]
\SetAlgoLined
\SetKwInOut{Input}{Input}
\SetKwInOut{Output}{Output}
\caption{Exhaustive Search}
\label{alg:exhaustive}
\Input{game-tree}
\Output{prompt}

opp-actions $\gets$ opponent's actions from game-tree\;
\For{action \KwSty{in} opp-actions}{
  leaves $\gets$ get leaves for action \label{line:leaf1}\;
  prompt $\gets$ action has expected-value for leaves\;
}
prompt $\gets$ compare actions to get best one\;

subtree $\gets$ game-tree if opponent takes best action\;

\For{action \KwSty{in} actions}{
  leaves $\gets$ get leaves for action in subtree \label{line:leaf2}\;
  prompt $\gets$ action has expected-value for leaves\;
}
prompt $\gets$ compare actions to get best response\;

\end{algorithm}
        \end{minipage}
        &
        \begin{minipage}[t]{0.47\textwidth}
            \begin{algorithm}[H]
\SetAlgoLined
\SetKwInOut{Input}{Input}
\SetKwInOut{Output}{Output}

\caption{Beliefs over hidden states}
\label{alg:bel1}

\Input{game-tree, opponent-action}
\Output{prompt}

hidden-states $\gets$ hidden in game-tree\;
\For{hidden-state \KwSty{in} hidden-states}{
  subtree $\gets$ get the subtree of hidden-state\;
  prompt $\gets$ add subtree of hidden-state\;
  value $\gets$ get expected value of opponent-action for subtree\;
  prompt $\gets$ hidden-state has value for opponent\;
}

prompt $\gets$ compare values of hidden-states to form belief\;

\end{algorithm}
        \end{minipage}
    \end{tabular}
\end{adjustbox}
\end{figure}

Search is conditioned on an opponent model, specifically, the levels of recursive reasoning (theory of mind)  that a player performs about other players' actions. So, a na\"ive  player $i$ -- a $0^{th}$ order theory of mind agent -- only searches through the rewards obtained for their own actions: $a_i^* = \text{argmax}_{a_i \in A_i}[\mathbb{E}_{s\in S}R_i(s, a_i)]$, assuming that the other players pick their action randomly; since the na\"ive player does not consider what other agents are thinking. A player $j$ that performs a higher level of reasoning (higher level theory of mind) first simulates the actions that the other player takes and then searches through their own actions to pick the best response $a_j^* = \text{argmax}_{a_j\in A_j}[\mathbb{E}_{s \in S} R(s, a_j, a_i^*)]$. We consider a na\"ive opponent model for the matrix games (though see \autoref{subsec:iterated} for extensions). 

As games increase in complexity, larger game trees and longer rule sets often surpass the context window of the LM, leading to less reliable reasoning. To address this, we propose \textbf{\textit{factoring}} reasoning via two tools \citep{schick2023toolformer,primer2022,nakano2021webgpt,dohan2022language}, `search' and `calculate'.
Search is a cognitive tool that dispatches to a new language model context with a specialized prompt. This allows the model to recursively search through subtrees of a game by generating ``\textit{search(agent, other agents, objective, action, other actions)}''. The calculate tools allow the model to call a simple calculator for either the `mean' or `argmax' operations, e.g.~via ``\textit{mean($a_1$=2, $a_2$=1)}''. In-context examples are adjusted to illustrate use of these tools (\autoref{asec:factor} for complete prompts and details).
Factoring in this way can enhance language models' capabilities to tackle complex games by enabling modular, effective search and reasoning.

When the space of  exhaustive search is too large, we carry out search with an iterative process of proposal and refinement of candidate actions \citep{morris2021generating,nelle2022evidence,primer2022,shinn2023reflexion}. In \autoref{alg:cand-search}, an action is proposed $a \in A$; the choice of this action reflects both priors learned by the model in pre-training and in-context examples. This action is evaluated, and revised to a new action to try and generate a better candidate action (this relies on in-context demonstrations to show action revision; (\autoref{alg:cand-search}: 1-4). This is repeated a predefined number of times which is specified in the demonstrations. For example, a step would look like: `` \textit{Try 1/N, If we play $a_1$, the expected reward would be $r_1$. This is low, so we can try again. Try 2/N, We revise our action to $a_2$. The reward would be $r_2$ ...}'' and so on (see sample prompt \autoref{prompt:broker}).

\begin{figure}[ht]
    \centering
    \begin{adjustbox}{max width=\textwidth}
    \begin{tabular}{@{}c@{\hspace{0.05\textwidth}}c@{}}
        \begin{minipage}[t]{0.5\textwidth}
\begin{algorithm}[H]
\SetAlgoLined
\caption{Search with proposals}
\label{alg:cand-search}

\SetKwInOut{Input}{Input}\SetKwInOut{Output}{Output}

\Input{game-tree, num-proposals}
\Output{prompt}

\For{attempt \KwSty{in} num-proposals}{
  action $\gets$ propose or revise action (game-tree)\;
  prompt $\gets$ action has expected-value(action)\;
}
prompt $\gets$ compare proposed actions to get best response\;
\end{algorithm}
        \end{minipage}
        &
        \begin{minipage}[t]{0.45\textwidth}
            \begin{algorithm}[H]
\SetAlgoLined
\SetKwInOut{Input}{Input}
\SetKwInOut{Output}{Output}

\caption{Evaluation}
\label{alg:val}

\Input{state/action, objective}
\Output{prompt}

leaf-nodes $\gets$ predict leaf nodes for state/ action\;
prompt $\gets$ action/state node has leaf-nodes\;
value, calculation-steps $\gets$ expected-value(leaf-nodes)\;
prompt $\gets$ action has value because calculation-steps\;

\end{algorithm}

        \end{minipage}
    \end{tabular}
\end{adjustbox}
\end{figure}

\textbf{Prompting Strategy: Value Assignment.} Understanding values flexibly is necessary for being able to generalize to new objectives. We alternate search with value assignment (\autoref{alg:exhaustive}: 3-4, 9-10); assigning values for each (state, action) pair: $R_i: S\times A_1\dots\times A_N \rightarrow \mathbb{R}$. We provide natural language explanations for the model to understand how values are calculated (\autoref{alg:val}: 3-4); for example, ``\textit{Bob wants to maximize his reward. As $a_1$ gives a higher reward compared to $a_2$, Bob will choose $a_1$}''. These explanations help in generalizing to a new value presented during the evaluation phase.

\textbf{Prompting Strategy: Belief Tracking.} We estimate beliefs about the hidden information such as values of players or non-observable states by searching through the possible space of hidden information (\autoref{fig:front}~b) and by keeping track of evidence in multiple interactions. For example, when the state $S$ is not visible but the other agents’ actions $A_i$ and rewards $R_i$ are observable, we search through the states and form a belief over them based on the actions and rewards of the other agents (see \autoref{alg:bel1} for pseudocode and \autoref{fig:front}~b for the demonstration). Similarly, if the values of other agents $R_i$ are hidden, we keep track of the actions to form a belief over what the rewards of an agent might be: ``\textit{Player chose the book over balls  $9/10$ times, so they must like books more.}'' (\autoref{fig:dealnodeal}~b).

\vspace{-3mm}
\section{Experiments}
\vspace{-3mm}
\label{sec:experiments}
Our experiments emphasize the utility of our prompting strategy by exploring the flexibility and reliability granted by its three main components: search \autoref{sec:search}, value assignemnt \autoref{sec:value}, and belief modelling \autoref{sec:beliefs}. See project website \footref{fn:web} for all prompts and evaluations.

\textbf{Language Models.} We evaluate a code-based language model, specifically \texttt{code-davinci-002} (CODEX \cite{chen2021evaluating}), and three text based models, \texttt{curie-001}, \texttt{davinci-002}, \texttt{davinci-003} \footnote{the rationale behind our selection of these Language Models (LMs) and results with text models can be found in \autoref{asec:lms}}. We sample greedily from the language model with the most deterministic setting, temperature 0. We allow the LM to sample freely during the ``reasoning'' phase; when it generates a stop phrase (``agent action:'') we then restrict the action space of the agent to valid actions using Synchromesh \citep{poesia2022synchromesh}. We convert the structure and rules of the games into text for the language model.

\textbf{Baselines.} We include the following baselines: 1) a 0-shot and a few-shot baseline where the model directly predicts the action without a scratchpad or chain-of-thought; 2) a 0-shot chain-of-thought baseline with a ``Let's think step by step:'' instruction \citep{kojima2022large}; 3) ablations where only specific components of the strategic prompts are used: only search, only value assignment, and search and value assignment together without belief modelling.

\textbf{Metrics.} We analyze performance by measuring the accuracy of choosing the best response. 

\vspace{-2mm}
\subsection{Searching with Language}
\vspace{-2mm}
\label{sec:search}
In this section, we show how language models guided with demonstrations generated from our method can learn to solve novel game trees reliably, generalizing to new rules, rewards, and structures. 

\begin{table}[!tbp]
    \centering
    \begin{adjustbox}{max width=\textwidth}
    \begin{tabular}{lcccccc}
    \toprule
       Games  & 0-shot & 0-shot CoT & 2-shot & Strategic & No Search & No Values \\
       \midrule
        Simultaneous 2x2  &0.29 (10/35)& 0.49 (17/35) & 0.37 (13/35)& \textbf{1.00} (35/35) &0.66 (23/35)& 0.88 (31/35)\\
        Sequential 2x2 & 0.60 (21/35)& 0.54 (19/35)& 0.71 (25/35)& \textbf{1.00} (35/35) & 0.74 (26/35) & 0.69 (24/35) \\

        2-stage game & 0.30 (9/30) & 0.37 (11/30) &  0.27 (8/30) & \textbf{0.90} (27/30) & 0.50 (15/30) & 0.47 (14/30) \\
    \bottomrule 
    \end{tabular}
    \end{adjustbox}
    \caption{Accuracy of predicting the best response in novel games for different prompt structures. We compare our strategic prompt with common baselines and ablations. Numbers in parentheses indicate number of correct choices out of the total number of evaluations.}
    \label{tab:search}
\end{table}

\vspace{-0.2cm}
\subsubsection{Matrix Games: New Rewards} \label{subsec:sim}
\textbf{Experiment Setup.} Here, we focus on testing generalization to new reward structures across different types of game structures; sequential games, simultaneours games and complex game formats involving both simultaneous and sequential play. 
The goal of the players in these games is to maximize their rewards by choosing optimal actions, either simultaneously or sequentially. We assume that the model is playing against a na\"ive (level-0) player (this is specified in the instruction). 

We evaluate language models using simple $2\times 2$ matrix games, focusing on single-stage simultaneous / sequential games and two-stage games. To teach strategic reasoning, we provide in-context examples of two basic matrix games with the appropriate format: First, we demonstrate a game with strictly descending payoffs ($8,7,\dots,1$) for all actions. Second, we demonstrate a game with payoff ties that matches the structure of the target evaluation game (simultaneous or sequential). These examples have the same game structure as, but simpler strategies than, that of the evaluation. These help teach the model to reason strategically before the main evaluation.
To evaluate generalization to new rewards, we test language models on different variations of matrix games: For single-stage games, we use 7 game types (Prisoner's Dilemma, Chicken, Stag Hunt, Battle of the Sexes, Market Entry, Imbalanced Matching Pennies, Deadlock; see \autoref{asec:22mat}) with 5 payoff variations each, yielding 35 games. These games are chosen as they require different types of behaviors to be successful: cooperation, compromise, and adversarial action. For 2-stage games, we use 2 game trees (simultaneous-sequential, sequential-simultaneous) with 15 payoff variations, yielding 30 games (\autoref{asec:complex}). The range of games evaluates their ability to show different strategic behaviors.

\textbf{Results.} Language models prompted to systematically search and assign values with our method generalize successfully to new payoffs (see \autoref{tab:search}). Based on the payoffs, models are able to express different types of behaviors. Vanilla strategies of prompting such as 0-shot, 2-shot, and 0-shot CoT prompts are less successful at generalizing to new payoffs. Further, when we ablate the strategic prompt to  remove parts of search or value assignment, performance becomes less reliable; showing how parts of our reasoning structure help in making LLMs more reliable.  

\vspace{-3mm}
\subsubsection{Complex Game Structures: Factoring Strategic Reasoning}
\vspace{-1mm}
\label{subsec:factor}
\begin{wrapfigure}[14]{r}{0.5\textwidth}
    \centering
 \includegraphics[width=\linewidth]{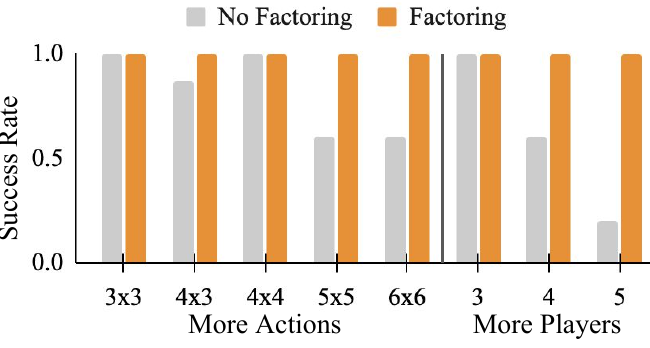}
    \caption{Comparison of accuracies of predicting best response with \& without factored reasoning.}
    \vspace{-3mm}
    \label{fig:factor}
\end{wrapfigure}

\textbf{Experiment Setup.} To further challenge language models, we evaluate their strategic reasoning on matrix games with increased complexity, without more complex in-context examples. We test games for generalization along two dimensions. First, larger action spaces, including  $3\times 3$, $4\times 3$, $5\times 5$, and $6 \times 6$ matrix games (sup. \autoref{afig:matrix-more}). As the action space grows, modeling the payoff structure and identifying optimal strategies becomes harder. Second, more players: 3, 4 and 5 player games.
Multi-player games introduce more complex dynamics including alliances, competition, and social dilemmas between groups. They require sophisticated reasoning about how different players may interact and influence each other. The payoffs for these games are extended versions of their $2\times2$ versions. For example, we use a 3-way prisoner's dilemma and chicken game to evaluate generalization to games with more players.
We still demonstrate strategic reasoning with only two trivial examples of solving $2\times 2$ matrix games. So, the model now has to generalize \textbf{\textit{0-shot}} to a new game structure and rewards. 

\textbf{Results.} As the size of the games increases, the language model is unable to keep up with the complexity of the problem, primarily due to its limited context size (see \autoref{fig:factor}). When we use factored search and reasoning, this is not a limitation and the model generalizes to the new games perfectly. We see that reasoning from in-context demonstrations allows models to flexibly generalize to games with more players and actions without any demonstrations for the new game structures. This result is promising as it shows that factoring allows us to generalize to much more complex games. We also show how factoring can potentially be used for iterated strategic reasoning in \autoref{subsec:iterated}.


\begin{table}[tpb]
    \centering
    \begin{adjustbox}{max width=\textwidth}
    \begin{tabular}{lcccccc}
    \toprule
       Objectives  & 0-shot & 0-shot CoT  & 2-shot & Strategic & No Search & No Value\\
       \midrule
        Max, Max & 0.29 (10/35)& 0.49 (17/35) &0.37 (13/35) & \textbf{1.0} (35/35)& 0.66 (23/35)& 0.88 (31/35)\\
        Max, Help  & 0.51 (18/35) &0.63 (22/35) & 0.49 (17/35)& \textbf{1.0} (35/35) &0.69 (24/35)&0.77 (27/35)\\
        Help, Max & 0.03 (1/35) & 0.51 (18/35)  & 0.43 (15/35) & \textbf{1.0} (35/35)&0.20 (7/35)& 0.63 (22/35)\\
        Welfare, Welfare & 0.63 (22/35) & 0.66 (23/35) &  0.60 (21/35) & \textbf{1.0} (35/35)&0.51 (18/35)&0.63 (22/35)\\
        Daxity, Daxity & 0.23 (8/35)& 0.43 (15/35)& 0.43 (15/35) & \textbf{0.94} (33/35)&0.60 (21/35) &0.71 (25/35)\\
    \bottomrule 
    \end{tabular}
    \end{adjustbox}
    \caption{Accuracy of choosing the best response for a new objective that has never been seen before. The goals in the first column indicate the objectives of the model and its opponent respectively; ``max, help'' means that the model is trying to maximize its payoff while its opponent is trying to help the model maximize the payoff. Daxity indicates a new objective that we define to be the difference between the payoff received by the player and its opponent.}
    \label{tab:value}
\end{table}
\vspace{-3mm}
\subsection{Assigning Values with Language}
\vspace{-2mm}
\label{sec:value}
In this section, we demonstrate how our method for evaluating states and actions in language can give large language models (LLMs) the flexibility to generalize and learn new objectives. We first show how our method can generalize to a new objective without any additional examples, then show how demonstrations can be used to teach the model to optimize for a new objective in a realistic setting.

\subsubsection{Matrix Games with Varying Objectives}

\textbf{Experiment Setup.} We consider four scenarios of objectives that the players seek to optimize: 1) the player tries to maximize its own reward knowing that the opponent is trying to help them; 2) the player tries to help the opponent maximize their reward given that the opponent is trying to maximize their reward; 3) Both players try to maximize the total welfare; and 4) The players maximize an arbitrary reward (`daxity'), defined as the advantage, which is the difference between their reward and the opponent's. We provide the same demonstrations as in section \autoref{subsec:sim}, where the players' objective is to maximize their reward. The model must generalize to the new objectives in a zero-shot manner; the model only sees demonstrations of players maximizing their payoffs and an instruction describing the new objective. The payoffs used to evaluate are the same as \autoref{subsec:sim}.

\textbf{Results.} Our reasoning structure with steps for inferring values in combination with search is able to generalize to alternate objectives (see \autoref{tab:value}). Here, language models can be seen as a powerful tool to optimize arbitrary new objectives. By generalizing to new objectives without re-training, we show that LMs can optimize not just for a fixed reward signal but for abstract goals. The ability to pursue new goals without losing competency is a hallmark of human-level reasoning, and success on this objective-generalization test is a promising sign, suggesting that LMs when prompted strategically, can show adaptable strategic reasoning.




\vspace{-2mm}
\subsubsection{Model as a fair broker}
\vspace{-8mm}
\begin{figure}[!tbhp]
    \centering
\begin{adjustbox}{max width=\textwidth}
    \begin{tabular}{@{}c@{\hspace{0.05\textwidth}}c@{}}
        \begin{minipage}[t]{0.6\textwidth}
            \vspace{0pt}
\begin{table}[H]
    \centering
    \begin{tabular}{cccc}
    \toprule
       Metrics & Strategic & No iterated search & Random\\
       \midrule
        Equality &\textbf{ 0.96} & 2.20 & 4.08\\
        Rawlsian & \textbf{1.20} & 2.61 & 4.06\\
    \bottomrule 
    \end{tabular}
    \caption{Difference from the optimal deal when the broker is optimizing for different fairness objectives. `Random' represents the expected fairness of a deal when a deal is randomly proposed.}
    \label{tab:broker}
\end{table}
        \end{minipage}
        &
        \begin{minipage}[t]{0.6\textwidth}
            \vspace{0pt}
\begin{table}[H]
    \centering
    \begin{tabular}{ccc}
    \toprule
       Games & Strategic & No Belief \\
       \midrule
        Beliefs: new payoffs & \textbf{1.0} (5/5) & 0.4 (2/5) \\
        Beliefs: more states & \textbf{1.0} (5/5) & 0.4 (2/5)\\
        Beliefs: more actions & \textbf{0.8} (4/5) & 0.6 (3/5)\\
        Beliefs: communication \autoref{asec:truth} &\textbf{ 1.00} (50/50)& 0.66 (33/50)\\
    \bottomrule 
    \end{tabular}
    \caption{Accuracy of predicting the best response when the model needs to form a belief over hidden information in the environment.}
    \label{tab:bel}
\end{table}
        \end{minipage}
    \end{tabular}
\end{adjustbox}

\end{figure}

\textbf{Experiment Setup.} In this section, we consider a realistic negotiation game where the model acts as a broker and has to optimize various fairness objectives for a successful deal. We consider the Deal or No Deal dataset \citep{lewis2017deal} to test the model. When the model acts as a broker, the model knows the values of each player and must propose a fair deal according to a fairness metric. The model is challenged to explore the complex action space of splitting the pot (where exhaustively searching all options is cumbersome) and adapting to new fairness objectives. We look at two different objectives that the broker has to optimize for: 1) equality: difference in the values that the players receive; 2) Rawlsian fairness: maximizing the minimum value that is given. 

We quantify performance by measuring the difference between the fairness of the proposed deal with the optimal deal. The model learns to optimize these objectives through few-shot demonstrations. We show 5 episodes of in-context demonstrations (see \autoref{asec:broker}) to teach the model to optimize an objective and teach it the reasoning structure of action proposal, evaluation, and revision. We test generalization to 100 new deals. The performance of the model is measured by the difference in the value of the proposed deal with the maximum possible value. We evaluate the model on 100 negotiation contexts from the Deal or No Deal dataset. 

\textbf{Results.} We see (\autoref{tab:broker}) that the iterated structure of reasoning enables the model to find more fair deals in comparison to a model without this capability, more than halving the error rate. This indicates the importance of iterative reasoning for negotiation models to converge on optimized solutions in complex search and value assignment problems.

\vspace{-3mm}
\subsection{Forming Beliefs with Language}
\vspace{-2mm}
\label{sec:beliefs}
\begin{figure*}[bt]
    \centering
\includegraphics[width=\textwidth]{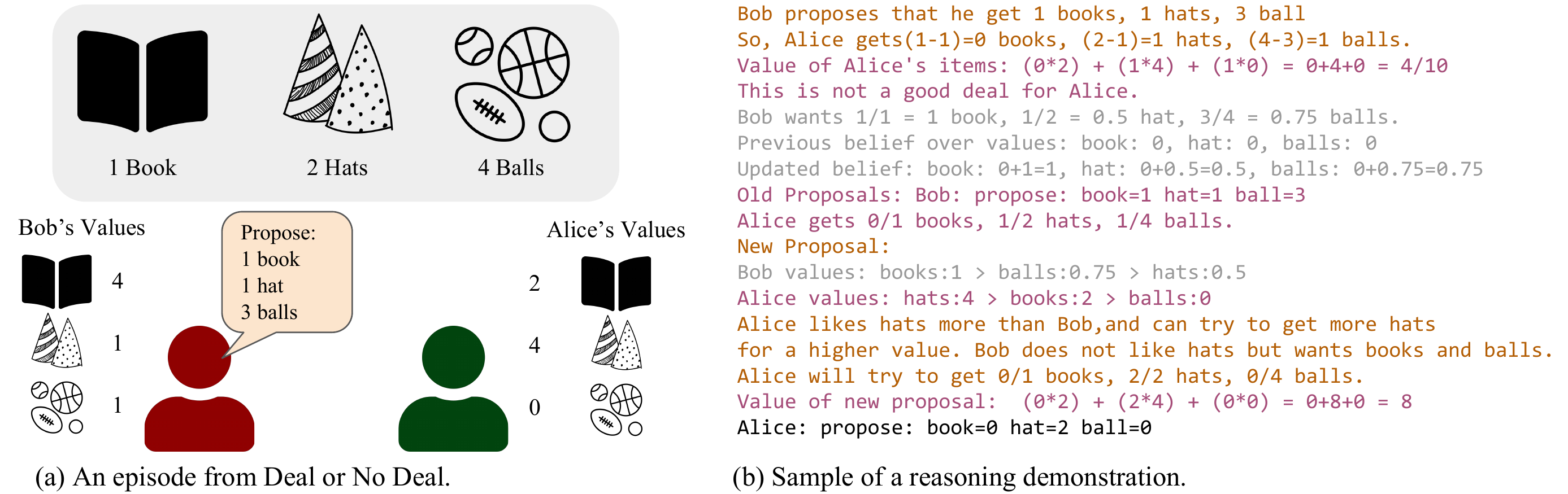}
    \caption{A sample episode (a) from Deal or No Deal \citep{lewis2017deal} with a (b) demonstration of the reasoning prompt; pink indicates reasoning over values, orange indicates search, and grey indicates beliefs.}
    \vspace{-2mm}
    \label{fig:dealnodeal}
\end{figure*}
 We show how LLMs prompted with our method to track beliefs can solve games with a variety of partially observable information, like hidden states, veracity of communication and opponent values.

\vspace{-2mm}
\subsubsection{Beliefs over States}
\vspace{-1mm}
\textbf{Experiment Setup.} In this section we introduce partial observability in the state of the world. We consider a sequential game where the player (model) cannot see the state of the world but can see the action that their opponent takes, and the payoffs that they receive for different world states. So, the player can observe the opponent's actions and values and must form a belief over the world state from these. We consider three types of game trees; (see \autoref{fig:front}) to show how language can be used for forming beliefs about hidden information. We show the prompt structure in \autoref{fig:front}. We evaluate their performance based on choosing the best response (response with the highest payoff).
To demonstrate reasoning over hidden states of the world, we use 2 examples of games with trivial payoffs (similar to previous sections). We use three game structures with five variations of payoffs to test generalization. The first game structure is the same as the in-context demonstrations but with non-trivial payoffs. In the second structure, the world has three hidden states instead of two. For the final structure, the number of actions that the opponent can take is increased from two to three.

\textbf{Results.} Models with our structures of reasoning outperform baselines, generalizing to new structures (\autoref{tab:bel}). This demonstrates how language enables models to reason under uncertainty and limited observability, a key capability for strategic reasoning and interaction.
The ability to form beliefs and reason under partial observability  suggests that language models can understand scenarios where information is imperfect and make inferences from limited signals. This type of flexible reasoning is crucial for interacting in complex, strategic environments. Our method shows one way of imparting this capability by using natural language to guide a model's search over possible hidden states. (Reasoning over the honesty of an opponent's communication, as in \autoref{asec:truth}, provides another example of how language can be used to form beliefs about hidden information based on limited signals.)

\begin{figure*}[!tbp]
    \centering
\includegraphics[width=\textwidth]{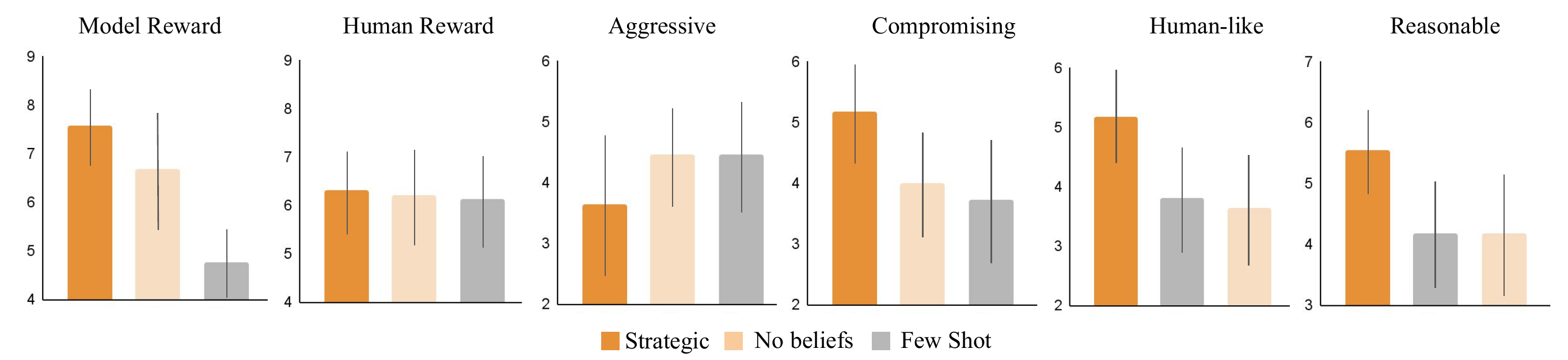}
    \caption{Comparison of different methods on different metrics when paired with real humans in the Deal or No Deal task. We show how our method outperforms baselines and ablations in human user studies. Error bars represent the standard deviation.}
    \vspace{-2mm}
    \label{fig:deal-res}
\end{figure*}





\vspace{-2.2mm}
\subsubsection{Realistic Scenario: Model as a negotiator}
\vspace{-1.23mm}

\textbf{Experiment Setup.} In this section, we show how we can create a human-like negotiation bot without retraining in a realistic negotiation setting by forming a belief over a human's values through multiple rounds of interaction. We use the Deal or No Deal environment \citep{lewis2017deal} to test our method (\autoref{fig:dealnodeal}). We restrict negotiations to a three turn (6 offers) setting. The human gets a chance to first propose an offer and then agents can propose a counter offer or accept/ reject the offer. When an agent rejects an offer, the player gets no reward. Demonstrations (\autoref{fig:dealnodeal}b) are provided by an expert human player collected while playing against another human. These are then annotated systematically by the player to generate 7 demonstrations. The demonstrations capture the style of the human player.

We evaluate the model on new negotiating partners in settings with new values and item quantities. We consider two baselines to compare our models against: 1) A few shot baseline without chain of thought prompting, 2) prompting with strategic reasoning that does not explicitly form beliefs over the values of the other player. We conducted a user study with 11 human participants that were asked to interact with the three negotiation methods for three episodes each. After interacting with the agents, the participants are asked to fill out a survey, rating the agent on different aspects of their negotiating ability on a 7-point Likert scale. Participants rated how human-like, reasonable, aggressive or compromising an agent was.

\textbf{Results.} Our method outperforms baselines on multiple metrics (\autoref{fig:deal-res}), achieving a higher average reward (7.6) compared to the `no belief' model (6.7) and the few-shot baseline (4.8). Despite our model getting a higher reward, the humans also received a slightly higher reward (6.30 vs 6.21) when negotiating with our method. We are successfully able to build a negotiating agent with no additional training, only through few-shot demonstrations. Further, the agent mimics the style of the human which is reflected in the post-study survey---our agent was consistently rated to be more human-like, reasonable, compromising and less aggressive. This shows how our method of prompting a model to reason strategically in language can enable agents to handle complex, real-world tasks, like negotiation, without requiring large training datasets. With improved reasoning and context, agents like this could have long, open interactions and even match human-level performance on social and strategic tasks. This initial result highlights the exciting potential for AI agents to collaborate with and assist people in new ways.

\vspace{-3mm}
\section{Discussion}
\vspace{-2mm}
\label{sec:discussion}
\vspace{-0.1cm}
In this paper, we set out to show how language models can be made to reason strategically about other agents by using a structured prompt based on search, value assignment and belief-tracking strategies. Our demonstrations enabled language models to generalize to strategic settings that had novel game rules and structures, new objectives and required forming beliefs over a variety of hidden information. These insights led to a model that negotiated with people in a considerate, human-like manner while maximizing utility. 

However, several challenges could arise in using language models for these domains. Complex games may require more demonstrations and instructions to understand the rules and game-specific behaviors; we are limited by the context length of the language model. Language models that rely on retrieval \citep{santhanam2021colbertv2,khattab2020colbert} could be used here, but extending them would be non-trivial. 
Complex games also require reasoning over longer time (i.e.~more turns) than considered here, which requires much longer reasoning traces where LM performance may degrade.
While our experiments on factored modeling are promising, other training strategies could help the model learn to tackle a new task quickly while maintaining flexibility by using language as a medium to "think."
For instance, supervised fine-tuning over reasoning trajectories, using rewards to tune the model \citep{bohm-etal-2019-better,stiennon2020learning}, or bootstrapping reasoning based on successful trajectories \citep{zelikman2022star}.  

We did not focus on finding equilibria, stable states of a game where no player has incentive to deviate from their strategy, in this work. However, the search and best-response strategies we proposed could potentially be extended to level-k reasoning, which converges on equilibrium. A simple extension of our factoring approach implements such recursive reasoning (see Appendix \ref{subsec:iterated} for an illustration).

We hope our results will inspire further research into using language models for reasoning in complex multi-agent settings such as Diplomacy and Hanabi \citep{paquette2019no,bard2020hanabi}; beyond merely being used for dialogue \citep{bakhtin2022human}. 
Our approach also has potential for modeling human-like behavior to form the basis for more realistic economic policy design, such as agent-based models of markets with human-like agents \citep{zheng2022econ}. These models could help simulate social conventions and agent-specific models of desires and beliefs, enabling more accurate simulations of complex social interactions. We hope our research will motivate further exploration into the use of language models for reasoning in such settings.


\section*{Acknowledgements}
This worked was supported by DARPA YFA, NSF Award \#2125511 and the NSF Expeditions Grant, Award Number (FAIN) 1918771. We would like to thank Minae Kwon, Gabriel Poesia, and Joseph Feffer for their feedback, insight, and discussions, which were instrumental in strengthening and improving this project.
\bibliography{main-ref}

\begin{thebibliography}{10}

\bibitem{aher2022using}
Gati Aher, Rosa~I Arriaga, and Adam~Tauman Kalai.
\newblock Using large language models to simulate multiple humans.
\newblock {\em arXiv preprint arXiv:2208.10264}, 2022.

\bibitem{bakhtin2022human}
Anton Bakhtin, Noam Brown, Emily Dinan, Gabriele Farina, Colin Flaherty, Daniel
  Fried, Andrew Goff, Jonathan Gray, Hengyuan Hu, Athul~Paul Jacob, Mojtaba
  Komeili, Karthik Konath, Minae Kwon, Adam Lerer, Mike Lewis, Alexander~H.
  Miller, Sasha Mitts, Adithya Renduchintala, Stephen Roller, Dirk Rowe, Weiyan
  Shi, Joe Spisak, Alexander Wei, David Wu, Hugh Zhang, and Markus Zijlstra.
\newblock Human-level play in the game of <i>diplomacy</i> by combining
  language models with strategic reasoning.
\newblock {\em Science}, 378(6624):1067--1074, 2022.

\bibitem{bakhtin2021no}
Anton Bakhtin, David Wu, Adam Lerer, and Noam Brown.
\newblock No-press diplomacy from scratch.
\newblock {\em Advances in Neural Information Processing Systems},
  34:18063--18074, 2021.

\bibitem{bard2020hanabi}
Nolan Bard, Jakob~N Foerster, Sarath Chandar, Neil Burch, Marc Lanctot,
  H~Francis Song, Emilio Parisotto, Vincent Dumoulin, Subhodeep Moitra, Edward
  Hughes, et~al.
\newblock The hanabi challenge: A new frontier for ai research.
\newblock {\em Artificial Intelligence}, 280:103216, 2020.

\bibitem{berner2019dota}
Christopher Berner, Greg Brockman, Brooke Chan, Vicki Cheung, Przemys{\l}aw
  D{\k{e}}biak, Christy Dennison, David Farhi, Quirin Fischer, Shariq Hashme,
  Chris Hesse, et~al.
\newblock Dota 2 with large scale deep reinforcement learning.
\newblock {\em arXiv preprint arXiv:1912.06680}, 2019.

\bibitem{bohm-etal-2019-better}
Florian B{\"o}hm, Yang Gao, Christian~M. Meyer, Ori Shapira, Ido Dagan, and
  Iryna Gurevych.
\newblock Better rewards yield better summaries: Learning to summarise without
  references.
\newblock In {\em Proceedings of the 2019 Conference on Empirical Methods in
  Natural Language Processing and the 9th International Joint Conference on
  Natural Language Processing (EMNLP-IJCNLP)}, pages 3110--3120, Hong Kong,
  China, November 2019. Association for Computational Linguistics.

\bibitem{brown2018superhuman}
Noam Brown and Tuomas Sandholm.
\newblock Superhuman ai for heads-up no-limit poker: Libratus beats top
  professionals.
\newblock {\em Science}, 359(6374):418--424, 2018.

\bibitem{chen2021evaluating}
Mark Chen, Jerry Tworek, Heewoo Jun, Qiming Yuan, Henrique Ponde de~Oliveira
  Pinto, Jared Kaplan, Harri Edwards, Yuri Burda, Nicholas Joseph, Greg
  Brockman, et~al.
\newblock Evaluating large language models trained on code.
\newblock {\em arXiv preprint arXiv:2107.03374}, 2021.

\bibitem{chowdhery2022palm}
Aakanksha Chowdhery, Sharan Narang, Jacob Devlin, Maarten Bosma, Gaurav Mishra,
  Adam Roberts, Paul Barham, Hyung~Won Chung, Charles Sutton, Sebastian
  Gehrmann, et~al.
\newblock Palm: Scaling language modeling with pathways.
\newblock {\em arXiv preprint arXiv:2204.02311}, 2022.

\bibitem{dohan2022language}
David Dohan, Winnie Xu, Aitor Lewkowycz, Jacob Austin, David Bieber,
  Raphael~Gontijo Lopes, Yuhuai Wu, Henryk Michalewski, Rif~A Saurous, Jascha
  Sohl-Dickstein, et~al.
\newblock Language model cascades.
\newblock {\em arXiv preprint arXiv:2207.10342}, 2022.

\bibitem{khattab2020colbert}
Omar Khattab and Matei Zaharia.
\newblock Colbert: Efficient and effective passage search via contextualized
  late interaction over bert.
\newblock In {\em Proceedings of the 43rd International ACM SIGIR conference on
  research and development in Information Retrieval}, pages 39--48, 2020.

\bibitem{kojima2022large}
Takeshi Kojima, Shixiang~Shane Gu, Machel Reid, Yutaka Matsuo, and Yusuke
  Iwasawa.
\newblock Large language models are zero-shot reasoners.
\newblock In Alice~H. Oh, Alekh Agarwal, Danielle Belgrave, and Kyunghyun Cho,
  editors, {\em Advances in Neural Information Processing Systems}, 2022.

\bibitem{kramar2022negotiation}
J{\'a}nos Kram{\'a}r, Tom Eccles, Ian Gemp, Andrea Tacchetti, Kevin~R McKee,
  Mateusz Malinowski, Thore Graepel, and Yoram Bachrach.
\newblock Negotiation and honesty in artificial intelligence methods for the
  board game of diplomacy.
\newblock {\em Nature Communications}, 13(1):1--15, 2022.

\bibitem{kwon2023reward}
Minae Kwon, Sang~Michael Xie, Kalesha Bullard, and Dorsa Sadigh.
\newblock Reward design with language models.
\newblock In {\em International Conference on Learning Representations (ICLR)},
  2023.

\bibitem{lake2017building}
Brenden~M Lake, Tomer~D Ullman, Joshua~B Tenenbaum, and Samuel~J Gershman.
\newblock Building machines that learn and think like people.
\newblock {\em Behavioral and brain sciences}, 40, 2017.

\bibitem{lewis2017deal}
Mike Lewis, Denis Yarats, Yann~N Dauphin, Devi Parikh, and Dhruv Batra.
\newblock Deal or no deal? end-to-end learning for negotiation dialogues.
\newblock In {\em Empirical Methods in Natural Language Processing (EMNLP)},
  2017.

\bibitem{moravvcik2017deepstack}
Matej Morav{\v{c}}{\'\i}k, Martin Schmid, Neil Burch, Viliam Lis{\`y}, Dustin
  Morrill, Nolan Bard, Trevor Davis, Kevin Waugh, Michael Johanson, and Michael
  Bowling.
\newblock Deepstack: Expert-level artificial intelligence in heads-up no-limit
  poker.
\newblock {\em Science}, 356(6337):508--513, 2017.

\bibitem{morris2021generating}
Adam Morris, Jonathan Phillips, Karen Huang, and Fiery Cushman.
\newblock Generating options and choosing between them depend on distinct forms
  of value representation.
\newblock {\em Psychological Science}, 32(11):1731--1746, 2021.

\bibitem{nakano2021webgpt}
Reiichiro Nakano, Jacob Hilton, Suchir Balaji, Jeff Wu, Long Ouyang, Christina
  Kim, Christopher Hesse, Shantanu Jain, Vineet Kosaraju, William Saunders,
  et~al.
\newblock Webgpt: Browser-assisted question-answering with human feedback.
\newblock {\em arXiv preprint arXiv:2112.09332}, 2021.

\bibitem{nelle2022evidence}
Jonas Nelle and Fiery Cushman.
\newblock Evidence for dynamic consideration set construction in open-ended
  problems.
\newblock In {\em Proceedings of the Annual Meeting of the Cognitive Science
  Society}, volume~44, 2022.

\bibitem{nye2022show}
Maxwell Nye, Anders~Johan Andreassen, Guy Gur-Ari, Henryk Michalewski, Jacob
  Austin, David Bieber, David Dohan, Aitor Lewkowycz, Maarten Bosma, David
  Luan, Charles Sutton, and Augustus Odena.
\newblock Show your work: Scratchpads for intermediate computation with
  language models.
\newblock In {\em Deep Learning for Code Workshop}, 2022.

\bibitem{openai-platform}
OpenAI.
\newblock Openai platform: Model index for researchers.
\newblock \url{https://platform.openai.com/docs/model-index-for-researchers}.
\newblock Accessed: May 8, 2023.

\bibitem{openai2023gpt4}
OpenAI.
\newblock Gpt-4 technical report, 2023.

\bibitem{ouyang2022training}
Long Ouyang, Jeff Wu, Xu~Jiang, Diogo Almeida, Carroll~L Wainwright, Pamela
  Mishkin, Chong Zhang, Sandhini Agarwal, Katarina Slama, Alex Ray, et~al.
\newblock Training language models to follow instructions with human feedback.
\newblock {\em arXiv preprint arXiv:2203.02155}, 2022.

\bibitem{paquette2019no}
Philip Paquette, Yuchen Lu, Seton~Steven Bocco, Max Smith, Satya O-G,
  Jonathan~K Kummerfeld, Joelle Pineau, Satinder Singh, and Aaron~C Courville.
\newblock No-press diplomacy: Modeling multi-agent gameplay.
\newblock {\em Advances in Neural Information Processing Systems}, 32, 2019.

\bibitem{poesia2022synchromesh}
Gabriel Poesia, Oleksandr Polozov, Vu~Le, Ashish Tiwari, Gustavo Soares,
  Christopher Meek, and Sumit Gulwani.
\newblock Synchromesh: Reliable code generation from pre-trained language
  models.
\newblock {\em arXiv preprint arXiv:2201.11227}, 2022.

\bibitem{santhanam2021colbertv2}
Keshav Santhanam, Omar Khattab, Jon Saad-Falcon, Christopher Potts, and Matei
  Zaharia.
\newblock Colbertv2: Effective and efficient retrieval via lightweight late
  interaction.
\newblock {\em arXiv preprint arXiv:2112.01488}, 2021.

\bibitem{sap2022neural}
Maarten Sap, Ronan Le~Bras, Daniel Fried, and Yejin Choi.
\newblock Neural theory-of-mind? on the limits of social intelligence in large
  lms.
\newblock In {\em EMNLP}, 2022.

\bibitem{schick2023toolformer}
Timo Schick, Jane Dwivedi-Yu, Roberto Dess{\`\i}, Roberta Raileanu, Maria
  Lomeli, Luke Zettlemoyer, Nicola Cancedda, and Thomas Scialom.
\newblock Toolformer: Language models can teach themselves to use tools.
\newblock {\em arXiv preprint arXiv:2302.04761}, 2023.

\bibitem{shinn2023reflexion}
Noah Shinn, Beck Labash, and Ashwin Gopinath.
\newblock Reflexion: an autonomous agent with dynamic memory and
  self-reflection.
\newblock {\em arXiv preprint arXiv:2303.11366}, 2023.

\bibitem{silver2016mastering}
David Silver, Aja Huang, Chris~J Maddison, Arthur Guez, Laurent Sifre, George
  Van Den~Driessche, Julian Schrittwieser, Ioannis Antonoglou, Veda
  Panneershelvam, Marc Lanctot, et~al.
\newblock Mastering the game of go with deep neural networks and tree search.
\newblock {\em nature}, 529(7587):484--489, 2016.

\bibitem{silver2018general}
David Silver, Thomas Hubert, Julian Schrittwieser, Ioannis Antonoglou, Matthew
  Lai, Arthur Guez, Marc Lanctot, Laurent Sifre, Dharshan Kumaran, Thore
  Graepel, et~al.
\newblock A general reinforcement learning algorithm that masters chess, shogi,
  and go through self-play.
\newblock {\em Science}, 362(6419):1140--1144, 2018.

\bibitem{stiennon2020learning}
Nisan Stiennon, Long Ouyang, Jeffrey Wu, Daniel Ziegler, Ryan Lowe, Chelsea
  Voss, Alec Radford, Dario Amodei, and Paul~F Christiano.
\newblock Learning to summarize with human feedback.
\newblock {\em Advances in Neural Information Processing Systems},
  33:3008--3021, 2020.

\bibitem{primer2022}
Andreas Stuhlmüller, Justin Reppert, and Luke Stebbing.
\newblock Factored cognition primer.
\newblock \url{https://primer.ought.org}, 2022.

\bibitem{suzgun2022challenging}
Mirac Suzgun, Nathan Scales, Nathanael Sch{\"a}rli, Sebastian Gehrmann, Yi~Tay,
  Hyung~Won Chung, Aakanksha Chowdhery, Quoc~V Le, Ed~H Chi, Denny Zhou, , and
  Jason Wei.
\newblock Challenging big-bench tasks and whether chain-of-thought can solve
  them.
\newblock {\em arXiv preprint arXiv:2210.09261}, 2022.

\bibitem{alpaca}
Rohan Taori, Ishaan Gulrajani, Tianyi Zhang, Yann Dubois, Xuechen Li, Carlos
  Guestrin, Percy Liang, and Tatsunori~B. Hashimoto.
\newblock Stanford alpaca: An instruction-following llama model.
\newblock \url{https://github.com/tatsu-lab/stanford_alpaca}, 2023.

\bibitem{valmeekam2022large}
Karthik Valmeekam, Alberto Olmo, Sarath Sreedharan, and Subbarao Kambhampati.
\newblock Large language models still can't plan (a benchmark for llms on
  planning and reasoning about change).
\newblock {\em arXiv preprint arXiv:2206.10498}, 2022.

\bibitem{vinyals2019grandmaster}
Oriol Vinyals, Igor Babuschkin, Wojciech~M Czarnecki, Micha{\"e}l Mathieu,
  Andrew Dudzik, Junyoung Chung, David~H Choi, Richard Powell, Timo Ewalds,
  Petko Georgiev, et~al.
\newblock Grandmaster level in starcraft ii using multi-agent reinforcement
  learning.
\newblock {\em Nature}, 575(7782):350--354, 2019.

\bibitem{wei2022chain}
Jason Wei, Xuezhi Wang, Dale Schuurmans, Maarten Bosma, brian ichter, Fei Xia,
  Ed~H. Chi, Quoc~V Le, and Denny Zhou.
\newblock Chain of thought prompting elicits reasoning in large language
  models.
\newblock In Alice~H. Oh, Alekh Agarwal, Danielle Belgrave, and Kyunghyun Cho,
  editors, {\em Advances in Neural Information Processing Systems}, 2022.

\bibitem{zelikman2022star}
Eric Zelikman, Yuhuai Wu, Jesse Mu, and Noah Goodman.
\newblock {ST}ar: Bootstrapping reasoning with reasoning.
\newblock In Alice~H. Oh, Alekh Agarwal, Danielle Belgrave, and Kyunghyun Cho,
  editors, {\em Advances in Neural Information Processing Systems}, 2022.

\bibitem{zheng2022econ}
Stephan Zheng, Alexander Trott, Sunil Srinivasa, David~C. Parkes, and Richard
  Socher.
\newblock The ai economist: Taxation policy design via two-level deep
  multiagent reinforcement learning.
\newblock {\em Science Advances}, 8(18):eabk2607, 2022.

\end{thebibliography}
\bibliographystyle{plain}
\clearpage
\appendix
\section{Additional Experiments}

\subsection{Search: 2x2 Sequential Matrix Games}
\label{subsec:seq}
\begin{figure}
    \centering
    \includegraphics[width=\textwidth]{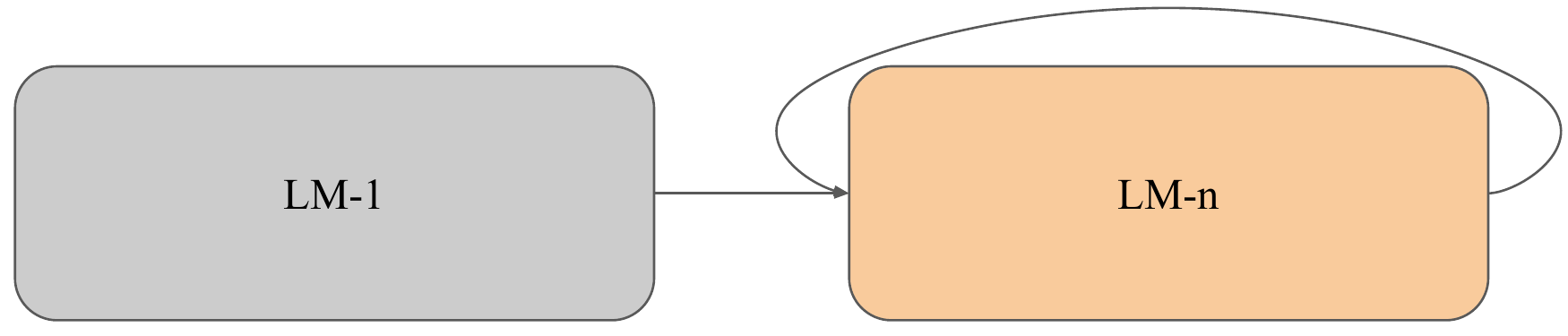}
    \caption{A visualization of an LM-cascade. The demonstrations for LM-1 and LM-n are different as the LM-n takes a prediction of an action as input.}
    \label{afig:casc}
\end{figure}

In this section, we consider games with sequential actions: the model plays an action, the other player observes this action and then takes an action with the highest payoffs. In this sequential setting, the first player has to search and reason through the opponent's actions, knowing that the opponent can see what they've played.

\textbf{In-Context Demonstrations.} The payoffs used for the in-context demos are the same as \autoref{subsec:sim} but the structure of reasoning is different as the actions are sequential.

\textbf{Evaluation Tasks.} We evaluate generalization to novel payoffs. We test on sequential versions of the games from \autoref{subsec:sim}.

\textbf{Results.} Similar to the previous section, we observe that prompts based on strategically searching and evaluating actions are more reliable than the baselines. The model is successful at taking the sequential form of play into account and choose the best response.

\subsection{Iterated Strategic Reasoning}
\label{subsec:iterated}
 \begin{table}[!hbt]
     \centering
     \begin{tabular}{lc}
     \toprule
        Method  &  Iterated Reasoning\\
        \midrule
         Level-1 Reasoner & 0.77 (28/35) \\
         No Cascade & 0.97 (34/35) \\
         Cascade & \textbf{1.00} (35/35) \\
         \bottomrule
     \end{tabular}
     \caption{Accuracy of predicting the best response when playing against a level-2 opponent.}
     \label{tab:iterated}
 \end{table}
We have so far only considered settings where the opponent was na\"ive and where the opponent did not think about the other players' strategies. Here, we remove this assumption and consider the case where the opponent knows that the player thinks about its action and chooses the best response according to this reasoning (level-2 reasoner). So, the player must iteratively find the best response to this opponent.

\textbf{Method.}  To incorporate n-iterations of strategic reasoning through the opponents' choices, we use a cascade or chain of language models (see \autoref{afig:casc}). For the first iteration, we assume that the opponent is na\"ive. The predicted action from this stage is passed on to a new language model context for search and reasoning to get the best response for the next level opponent: `\textit{`I thought through my opponent's actions and chose action 1. But if my opponent knows that I thought through their actions and that I will play action 1, then ...}''. 

\textbf{In-Context Demonstrations.}  We use the same in-context demonstrations from \autoref{subsec:sim} for the level-1 model, and provide two demonstrations on trivial problems for the level-n model. The demonstrations for the level-n model take into account the action predicted by the model from the previous level. We consider a level-2 reasoner via both a LM cascade and  by combining the reasoning prompts for the level-1 and level-2 model in a single context.

\textbf{Evaluation Tasks.} We consider $2\times 2$ matrix games from \autoref{subsec:sim} but the player plays against a level-2 opponent. 

\textbf{Results.} The cascade generalizes successfully to a level-2 opponent (\autoref{tab:iterated}). This shows that our prompting strategy combined with the modularity of an LLM-cascade is a powerful tool for an arbitrary level of iterated strategic reasoning. The no-cascade baseline also performs well, though we expect performance would degrade for higher levels of reasoning. The level-1 reasoner is only effective in games where the optimal strategy is the same as that against a level-0 opponent.

\subsection{Beliefs of Truthfulness}
\label{asec:truth}
Here, we consider games where the opponent can communicate with the player and tell them what action they are going to play. The opponent can be faithful to what they say they are going to play or lie about their intentions. The player must infer if their opponent is telling the truth or is lying based on the payoffs, and then choose the best response to their opponents actions. We show how language can help understand and reason over the veracity of communication. Veracity can be seen as a hidden state of the world (the opponent) that the model must infer from the evidence. We use a reasoning structure similar to the previous section. The opponent in our evaluations is na\"ive, and assumes that the player will believe whatever they are told.

\textbf{In-Context Demonstrations.} We use one demonstration with trivial payoffs to teach the model to form a belief over the honesty of the communication. In this demonstration, the opponent says the truth. We use a second demonstration with trivial payoffs to teach the model how to break ties (same as \autoref{subsec:sim}).

\textbf{Evaluation Tasks.} We test the model on 5 types of $2\times 2$ matrix games (cooperation, matching pennies, prisoner's dilemma, chicken, battle of the sexes; see \autoref{asec:matrix}) with 5 variations of payoffs. These games are designed such that some require deception while others rely on the honesty of communication.

\begin{table}[H]
    \centering
    \begin{tabular}{ccc}
    \toprule
       Games & Strategic & No Belief \\
       \midrule
        Beliefs: communication \autoref{asec:truth} &\textbf{ 1.00} (50/50)& 0.66 (33/50)\\
    \bottomrule 
    \end{tabular}
    \caption{Accuracy of predicting the best response when the model needs to form a belief over hidden information in the environment.}
    \label{atab:truth}
\end{table}

\textbf{Results.} We find that models that search through different states of the world based on the payoffs and actions are able to infer the veracity of the communication reliably. The model is successful in generalizing to new payoff structures and to scenarios where the opponent lies about their intentions. In general, this section shows how language can be used to form beliefs over hidden mental states.

\subsection{Evaluations with other LMs}
\label{asec:lms}
\textbf{Justification for choosing \texttt{code-davinci-002.}} We chose a model which is state of the art for several long-form reasoning problems \citep{suzgun2022challenging, wei2022chain} other than GPT-4 \citep{openai2023gpt4} and PALM \citep{chowdhery2022palm}. Moreover, it is also the best performing model (except for PALM, GPT-4) on mathematics and coding domains; both of which require long-form compositional reasoning similar to our problems \citep{wei2022chain}. Anecdotally, \texttt{code-davinci-002} was also more consistent at reasoning compared to \texttt{text-davinci-002} in our preliminary explorations.

\textbf{Experiments with Other LMs.} To ensure that our prompts hold for other language models, we have run three additional models on our evaluations for search and value assignment. We used \texttt{text-curie-001} as a small instruction-tuned baseline, \texttt{text-davinci-002} as an instruction-tuned text model (which uses \texttt{code-davinci-002} as a base-model)\citep{openai-platform}, and \texttt{text-davinci-003} which is an RLHF-tuned version of \texttt{text-davinci-002}. See the results of the models in \autoref{tab:lms-search} and in \autoref{tab:lms-val}.

\begin{table}[!htbp]
    \centering
\begin{adjustbox}{max width=\textwidth}
\begin{tabular}{lcccc}
\toprule
& Simultaneous 2x2 & More actions 3x3, 4x3 & More players 2x2x2 \\
\midrule
\texttt{code-davinci-002}: 2-shot & 0.37 (13/35) & 0.43 (13/30) & 0.20 (3/15) \\
\texttt{code-davinci-002}: Strategic & 1.00 (35/35) & 0.93 (28/30) & 1.00 (15/15) \\
\texttt{text-davinci-003}: 2-shot & 0.66 (23/35) & 0.43 (13/30) & 0.13 (2/15) \\
\texttt{text-davinci-003}: Strategic & 1.00 (35/35) & 0.66 (23/35) & 0.66 (10/15) \\
\texttt{text-davinci-002}: 2-shot & 0.49 (17/35) & 0.40 (12/30) & 0.13 (2/15) \\
\texttt{text-davinci-002}: Strategic & 0.80 (28/35) & 0.70 (21/30) & 0.27 (4/15) \\
\texttt{text-curie-001}: 2-shot & 0.77 (27/35) & 0.40 (12/30) & 0.47 (7/15) \\
\texttt{text-curie-001}: Strategic & 0.77 (27/35) & - & - \\
\bottomrule
\end{tabular}
\end{adjustbox}
    \caption{Results for search with different language models.}
    \label{tab:lms-search}
\end{table}

\begin{table}[!htbp]
    \centering
\begin{adjustbox}{max width=\textwidth}
\begin{tabular}{lccccc}
\toprule
& max,max & max,help & help,max & welfare,welfare & daxity,daxity \\
\midrule
\texttt{code-davinci-002}: 2-shot & 0.37 (13/35) & 0.49 (17/35) & 0.43 (15/35) & 0.60 (21/35) & 0.43 (15/35) \\
\texttt{code-davinci-002}: Strategic & 1.00 (35/35) & 1.00 (35/35) & 1.00 (35/35) & 1.00 (35/35) & 0.94 (33/35) \\
\texttt{text-davinci-003}: 2-shot & 0.66 (23/35) & 0.60 (21/35) & 0.49 (17/35) & 0.74 (26/35) & 0.71 (25/35) \\
\texttt{text-davinci-003}: Strategic & 1.00 (35/35) & 1.00 (35/35) & 0.86 (30/35) & 0.94 (33/35) & 0.89 (31/35) \\
\texttt{text-davinci-002}: 2-shot & 0.49 (17/35) & 0.69 (24/35) & 0.43 (15/35) & 0.69 (24/35) & 0.83 (29/35) \\
\texttt{text-davinci-002}: Strategic & 0.80 (28/35) & 1.00 (35/35) & 0.74 (26/35) & 0.80 (28/35) & 0.80 (28/35) \\
\texttt{text-curie-001}: 2-shot & 0.77 (27/35) & 0.43 (15/35) & 0.49 (17/35) & 0.60 (21/35) & 0.74 (26/35) \\
\texttt{text-curie-001}: Strategic & 0.77 (27/35) & 0.46 (16/35) & 0.46 (16/35) & 0.67 (23/35) & 0.57 (20/35) \\
\bottomrule
\end{tabular}
\end{adjustbox}
    \caption{Results for value assignment with different language models.}
    \label{tab:lms-val}
\end{table}

\section{Demonstration Prompts}
\label{asec:prompts}
\begin{table}[!htbp]
    \centering
    \begin{tabular}{ll}
    \toprule
    Prompt & \# Tokens \\
    \midrule
4.1.1 & 1423 \\
4.1.2 & 2292 (seq+sim), 2316 (sim+seq) \\
4.1.3 & 1423 (level 1), 1280 (level-n) \\
4.2.1 & 1423 \\
4.2.2 & 4054 (equality), 2455 (rawlsian) \\
4.3.1 & 2640 \\
4.3.2 & 1920 \\
4.3.3 & 4268 \\
\bottomrule
    \end{tabular}
    \caption{Token statistics for prompts used in different experiments. Tokens are counted based on the \texttt{code-davinci-002} tokenizer.}
    \label{tab:tokens}
\end{table}

\textbf{Reproducibility} We provide the prompts that are used to teach the models. Full prompts used in our experiments are available at the project homepage\footnote{Prompts are available here \url{https://sites.google.com/view/strategic-reasoning-llms/}}. Token statistics for the prompts are shown in \autoref{tab:tokens}. The models we test are hidden behind a paid API and \texttt{code-davinci-002} is now available only as part of a researcher access program. The parameters of these models, the specifics of training these models, and the data used to train these models are not publicly accessible.

We include snippets of demonstrations here.

\subsection{2x2 Matrix Games}
\label{asec:22mat}
See \autoref{prompt:mat} for one demonstration from our prompt.
\begin{figure*}[hbtp]
    \caption{One demonstration of how to solve a trivial $2\times 2$ matrix game.}
    \label{prompt:mat}
\begin{tcolorbox}[
    prompt,
    title={\small \textbf{One demonstration of how to solve a trivial $2\times 2$ matrix game.}},
    width=\textwidth]
    \fontsize{7pt}{7pt}\selectfont
    Q:Gopher and Bob are playing a game. Gopher and Bob get rewards according to these rules:\\
1. Gopher:a1, Bob:b1, then Gopher reward gr11=8 and Bob reward br11=7\\
2. Gopher:a1, Bob:b2, then Gopher reward gr12=6 and Bob reward br12=5\\
3. Gopher:a2, Bob:b1, then Gopher reward gr21=4 and Bob reward br21=3\\
4. Gopher:a2, Bob:b2, then Gopher reward gr22=2 and Bob reward br22=1\\
Gopher and Bob are trying to maximize their reward. What action should Gopher play?\\\\
A:Let's reason about what Bob wants to first.\\
Bob wants to maximize his reward.\\
If Bob plays b1,\\
1. Gopher:a1, Bob:b1, Gopher reward gr11=8 and Bob reward br11=7: Bob's reward r11=(br11)=br11=7 \\
3. Gopher:a2, Bob:b1, Gopher reward gr21=4 and Bob reward br21=3: Bob's reward r21=(br21)=br21=3\\
So, Bob's expected reward for b1 is (r11+r21)/2 = (7+3)/2 = 10/2 = 5\\
If Bob plays b2,\\
2. Gopher:a1, Bob:b2, then Gopher reward gr12=6 and Bob reward br12=5: Bob's reward r12=(br12)=br12=5\\
4. Gopher:a2, Bob:b2, then Gopher reward gr22=2 and Bob reward br22=1: Bob's reward r22=(br22)=br22=1\\
So, Bob's expected reward for b2 is (r12+r22)/2 = (5+1)/2 = 6/2 = 3\\
As b1=5, b2=3, 5>3, b1>b2, Bob will play b1.\\
Now let's reason for Gopher.\\
Gopher wants to maximize its reward.\\
As we know Bob plays b1 we dont need to calculate expected reward for each action, listing actions where Bob plays b1 is enough.\\
1. Gopher:a1, Bob:b1, then Gopher reward gr11=8 and Bob reward br11=7: Gopher's reward r11=(gr11)=gr11=8\\
3. Gopher:a2, Bob:b1, then Gopher reward gr21=4 and Bob reward br21=3: Gopher's reward r21=(gr21)=gr21=4\\
We know Bob plays b1, so Gopher will pick the action with the highest payoff.\\
As a1=8, a2=4, 8>4, a1>a2, Gopher will play a1.\\
Gopher's action:a1
\normalfont
\end{tcolorbox}
\end{figure*}

\subsection{Factoring Reasoning}
\label{asec:factor}
We have two prompts for factoring reasoning: the recursive case and the base case. The base case is shown in \autoref{prompt:rec} and the base case is shown in \autoref{prompt:base}.

\begin{figure}[hbtp]
\centering
\begin{adjustbox}{max width=3\textwidth}
    \caption{Prompts for factored reasoning. Demonstrations to solve trivial $2\times 2$ matrix games.}
    \label{prompt:rec}
\begin{tcolorbox}[
    prompt,
    title={\small \textbf{Prompts for factored reasoning. Demonstrations to solve trivial $2\times 2$ matrix games.}},
    width=\textwidth]
    \fontsize{7pt}{7pt}\selectfont
There are 3 functions that help in solving the problems: 

1. search: search(agent, other\_agent, objective, action, other\_actions) returns the expected reward for the agent. other actions are passed when the agent thinks about other agents.

2. compare: compare(agent, objective, [list of actions]) returns the action that maximizes the agent's objective.

3. mean: mean([list of numbers]) returns the mean of the list of numbers.
\\ \\
Q:Gopher and Bob are playing a game. Gopher and Bob get rewards according to these rules:

1. Gopher:a1, Bob:b1, then Gopher reward gr11=-3 and Bob reward br11=-2

2. Gopher:a1, Bob:b2, then Gopher reward gr12=-1 and Bob reward br12=-4

3. Gopher:a2, Bob:b1, then Gopher reward gr21=1 and Bob reward br21=2

4. Gopher:a2, Bob:b2, then Gopher reward gr22=3 and Bob reward br22=4

Gopher and Bob are trying to maximize their reward. Bob does not think about their opponent and only plays the action with the highest expected reward. Gopher thinks about other players' reasoning. What action should Gopher play? Pick from a1, a2, a1 or a2?
\\ \\
A:Let's reason about what Bob wants to first.

Bob wants to maximize his reward: br.

If Bob plays b1, exepected reward for b1 is search(Bob, Gopher, max, b1) = 0.

If Bob plays b2, exepected reward for b2 is search(Bob, Gopher, max, b2) = 0.

So, Bob will play compare(Bob, max, [b1=0, b2=0])= [b1,b2].

Now let's reason for Gopher.

Gopher wants to maximize its reward: gr.

As Bob plays b1 or b2 we calculate the expected reward for each action,

If Gopher plays a1, exepected reward for a1 is search(Gopher, Bob, max, a1, [bob[b1, b2]]) = -2.

If Gopher plays a2, exepected reward for a2 is search(Gopher, Bob, max, a2, [bob[b1, b2]]) = 2.

So, Gopher will play compare(Gopher, max, [a1=-2, a2=2]) = [a2].

Gopher's action:a2. \\\\
Q:Gopher and Bob are playing a game. Gopher and Bob get rewards according to these rules:

1. Gopher:a1, Bob:b1, then Gopher reward gr11=8 and Bob reward br11=7

2. Gopher:a1, Bob:b2, then Gopher reward gr12=6 and Bob reward br12=5

3. Gopher:a2, Bob:b1, then Gopher reward gr21=4 and Bob reward br21=3

4. Gopher:a2, Bob:b2, then Gopher reward gr22=2 and Bob reward br22=1

Gopher and Bob are trying to maximize their reward. Bob does not think about their opponent and only plays the action with the highest expected reward. Gopher thinks about other players' reasoning. What action should Gopher play? Pick from a1, a2, a1 or a2?\\

A:Let's reason about what Bob wants to first.
Bob wants to maximize his reward: br.

If Bob plays b1, exepected reward for b1 is search(Bob, Gopher, max, b1) = 5.

If Bob plays b2, exepected reward for b2 is search(Bob, Gopher, max, b2) = 3.

So, Bob will play compare(Bob, max, [b1=5, b2=3]) = [b1].

Now let's reason for Gopher.

Gopher wants to maximize its reward: gr.

As Bob plays b1, we calculate the expected reward for each action,

If Gopher plays a1, exepected reward for a1 is search(Gopher, Bob, max, a1, [bob[b1]]) = 8.

If Gopher plays a2, exepected reward for a2 is search(Gopher, Bob, max, a2, [bob[b1]]) = 4.

So, Gopher will play compare(Gopher, max, [a1=8, a2=4]) = [a1].

Gopher's action:a1.
\normalfont
\end{tcolorbox}
\end{adjustbox}
\end{figure}
\begin{figure*}[hbtp]
\centering
\caption{One demonstration of how to solve a trivial $2\times 2$ matrix game.}
\label{prompt:base}
\begin{tcolorbox}[
    prompt,
    title={\small \textbf{One demonstration of how to solve a trivial $2\times 2$ matrix game.}},
    width=\textwidth]
    \fontsize{7pt}{7pt}\selectfont
Q:Gopher and Bob are playing a game. Gopher and Bob get rewards according to these rules:

1. Gopher:a1, Bob:b1, then Gopher reward gr11=8 and Bob reward br11=7

2. Gopher:a2, Bob:b1, then Gopher reward gr12=6 and Bob reward br12=5

Bob's objective: max. What is Bob's expected reward?

A: Bob is maximizing his reward: br.

If Gopher plays a1,

1. Gopher:a1, Bob:b1, then Gopher reward gr11=8 and Bob reward br11=7; Bob maximizes r11=(br11)=br11=7

If Gopher plays a2,

2. Gopher:a2, Bob:b1, then Gopher reward gr21=4 and Bob reward br21=3; Bob maximizes r21=(br21)=br21=3

Expected reward for Bob = mean([r11, r21]) = mean([7, 3]) = 5.

Answer:5.

Q:Gopher and Bob are playing a game. Gopher and Bob get rewards according to these rules:

1. Gopher:a1, Bob:b1, then Gopher reward gr11=8 and Bob reward br11=7

2. Gopher:a1, Bob:b2, then Gopher reward gr12=6 and Bob reward br12=5

Gopher's objective: max. What is Gopher's expected reward?

A: Gopher is maximizing its reward: gr.

If Bob plays b1,

1. Gopher:a1, Bob:b1, then Gopher reward gr11=8 and Bob reward br11=7; Gopher maximizes r11=(gr11)=gr11=8

If Bob plays b2,

2. Gopher:a1, Bob:b2, then Gopher reward gr12=6 and Bob reward br12=5; Gopher maximizes r12=(gr12)=gr12=6

Expected reward for Gopher = mean([r11, r12]) = mean([8, 6]) = 7.

Answer:7.
\normalfont
\end{tcolorbox}
\end{figure*}

\subsection{Deal or No Deal: Broker}
\label{asec:broker}
See \autoref{prompt:broker} for one demonstration from our prompt for optimizing equality.

\begin{figure*}[hbtp]
\vspace{-1.5cm}
    \caption{One demonstration of optimizing for equality as a broker in negotiation.}
    \label{prompt:broker}
\begin{tcolorbox}[
    prompt,
    title={\small \textbf{One demonstration of optimizing for equality as a broker in negotiation.}},
    width=\textwidth]
    \fontsize{7pt}{7pt}\selectfont
Alice and Bob are trying to split books, hats and balls amongst themselves. The value of the books, hat and balls changes across scenarios. The items have a different value for Alice and Bob. Propose a deal where the Alice and Bob receive equal/ similar rewards. Try proposing 3 times at most.\\\\

\#\# New Deal\\
items available:\\
book=3\\
hat=1\\
ball=2\\
Bob's values: book=0 hat=2 ball=4\\
Alice's values: book=1 hat=3 ball=2\\
Q: What is a proposal that gives similar payoffs?\\\\
Try 1/3.\\
items: book=3, hat=1, ball=2\\
Bob values: ball:4 > hat:2 > book:0\\
Alice values: hat:3 > ball:2 > book:1\\
In a proposal with similar payoffs, Alice gets 3 books, 1 hats, 0 balls.\\
So, Bob gets (3-3)=0 books, (1-1)=0 hats, (2-0)=2 balls.\\
Value of proposal for Alice:  (3*1) + (1*3) + (0*2) = 3+3+0 = 6/10\\
Value of proposal for Bob:  (0*0) + (0*2) + (2*4) = 0+0+8 = 8/10\\
Alice gets 6/10 and Bob gets 8/10. Difference in payoffs 8-6 = 2. This is a good deal but we might be able to do better.\\
So, let's try again.\\\\
Try 2/3.\\
items: book=3, hat=1, ball=2\\
Bob values: ball:4 > hat:2 > book:0\\
Alice values: hat:3 > ball:2 > book:1\\
old proposal: Alice gets 3 books, 1 hats, 0 balls. Alice gets 6/10 and Bob gets 8/10.\\
new proposal: Alice gets 3 books, 0 hats, 1 balls.\\
So, Bob gets (3-3)=0 books, (1-0)=1 hats, (2-1)=1 balls.\\
Value of proposal for Alice:  (3*1) + (0*3) + (1*2) = 3+0+2 = 5/10\\
Value of proposal for Bob:  (0*0) + (1*2) + (1*4) = 0+2+4 = 6/10\\
Alice gets 5/10 and Bob gets 6/10. Difference in payoffs 6-5 = 1. This is a good deal but we might be able to do better.\\
So, let's try again.\\
Try 3/3. last try.\\\\
items: book=3, hat=1, ball=2\\
Bob values: ball:4 > hat:2 > book:0\\
Alice values: hat:3 > ball:2 > book:1\\
old proposal: Alice gets 3 books, 0 hats, 1 balls. Alice gets 5/10 and Bob gets 6/10.\\
new proposal: Alice gets 2 books, 0 hats, 2 balls.\\
So, Bob gets (3-2)=1 books, (1-0)=1 hats, (2-2)=0 balls.\\
Value of proposal for Alice:  (2*1) + (0*3) + (2*2) = 2+0+4 = 6/10\\
Value of proposal for Bob:  (1*0) + (1*2) + (0*4) = 0+2+0 = 2/10\\
Alice gets 6/10 and Bob gets 2/10. Difference in payoffs 6-2 = 4. This is a bad deal as payoffs are not close.\\
Summary of tries:\\
Try 1/3: Alice gets 6/10 and Bob gets 8/10. Difference in payoffs 8-6 = 2.\\
Try 2/3: Alice gets 5/10 and Bob gets 6/10. Difference in payoffs 6-5 = 1.\\
Try 3/3: Alice gets 6/10 and Bob gets 2/10. Difference in payoffs 6-2 = 4.\\
So, the best deal is the second try.\\
Minimum difference is 2. So, we go with try 2/3. Alice gets 0 books, 1 hats, 2 balls.\\
propose: book=0 hat=1 ball=2
\normalfont
\end{tcolorbox}
\end{figure*}
\subsection{Deal or No Deal: Player}
See \autoref{prompt:neg} for an expert annotated demonstration of a negotiation.
\begin{figure*}[hbtp]
\vspace{-1cm}
    \caption{A demonstration of negotiating a deal with an agent.}
    \label{prompt:neg}
    \begin{tcolorbox}[
    prompt,
    title={\small \textbf{A demonstration of negotiating a deal with an agent.}},
    width=\textwidth]
    \fontsize{7pt}{7pt}\selectfont
\#\# New Deal\\
items available:\\
book=1\\
hat=4\\
ball=1\\
Alice's values: book=4 hat=1 ball=2\\
prior over Bob's values: book=0 hat=0 ball=0\\\\
Bob: propose: book=0 hat=3 ball=1\\\\
Let's think step by step for Alice:\\
Bob proposes that he get 0 books, 3 hats, 1 ball\\
So, Alice gets(1-0)=1 books, (4-3)=1 hats, (1-1)=0 balls.\\
Value of Alice's items: (1*4) + (1*1) + (0*2) = 4+1+0 = 5/10\\
This is not a good deal for Alice.\\
Bob wants 0/1 = 0 book, 3/4 = 0.75 hat, 1/1 = 1 balls.\\
Previous belief over values: book: 0, hat: 0, balls: 0\\
Updated belief: book: 0+0=0, hat: 0+0.75=0.75, balls: 0+1=1\\
Old Proposals:\\
Bob: propose: book=0 hat=3 ball=1\\
Alice gets 1/1 books, 1/4 hats, 0/1 balls.\\
New Proposal that is different from old proposals:\\
Bob values: balls:1 > hats:0.75 > books:0\\
Alice values: books:4 > balls:2 > hats:1\\
Alice like books more than Bob, and can try to get more hats for a higher value. Bob does not like books but wants balls.\\
Alice will try to get 1/1 books, 4/4 hats, 0/1 balls.\\
Value of new proposal:  (1*4) + (4*1) + (0*2) = 4+4+0 = 8\\
Alice: propose: book=1 hat=4 ball=0\\\\

Bob: propose: book=0 hat=2 ball=1\\\\

Let's think step by step for Alice:\\
Bob proposes that he get 0 books, 2 hats, 1 ball\\
So, Alice gets (1-0)=1 books, (4-2)=2 hats, (1-1)=0 balls.\\
Value of Alice's items: (1*4) + (2*1) + (0*2) = 4+2+0 = 6/10\\
This is not a great deal for Alice.\\
Bob wants 0/1 = 0 book, 2/4 = 0.5 hat, 1/1 = 1 balls.\\
Previous belief over values: book: 0, hat: 0.75, balls: 1\\
Updated belief: book: 0+0=0, hat: 0.75+0.5=1.25, balls: 1+1=2\\
Old Proposals:\\
Alice: propose: book=1 hat=4 ball=0\\
Alice gets 1/1 books, 4/4 hats, 0/1 balls.\\
Bob: propose: book=0 hat=2 ball=1\\
Alice gets 1/1 books, 2/4 hats, 0/1 balls.\\
New Proposal that is different from old proposals:\\
Bob values: balls:2 > hats:1.25 > books:0\\
Alice values: books:4 > balls:2 > hats:1\\
Alice can give take one more hat to increase her value. Bob likes balls, doesn't want books.\\
Alice will try to get 1/1 books, 3/4 hats, 0/1 balls.\\
Value of new proposal:  (1*4) + (3*1) + (0*2) = 4+3+0 = 7/10\\
Alice: propose: book=1 hat=3 ball=0\\\\

Bob:accept
\normalfont
\end{tcolorbox}
\end{figure*}


\section{Evaluation Tasks}
\subsection{Matrix Games}
\begin{figure*}
    \centering
    \includegraphics[width=\textwidth]{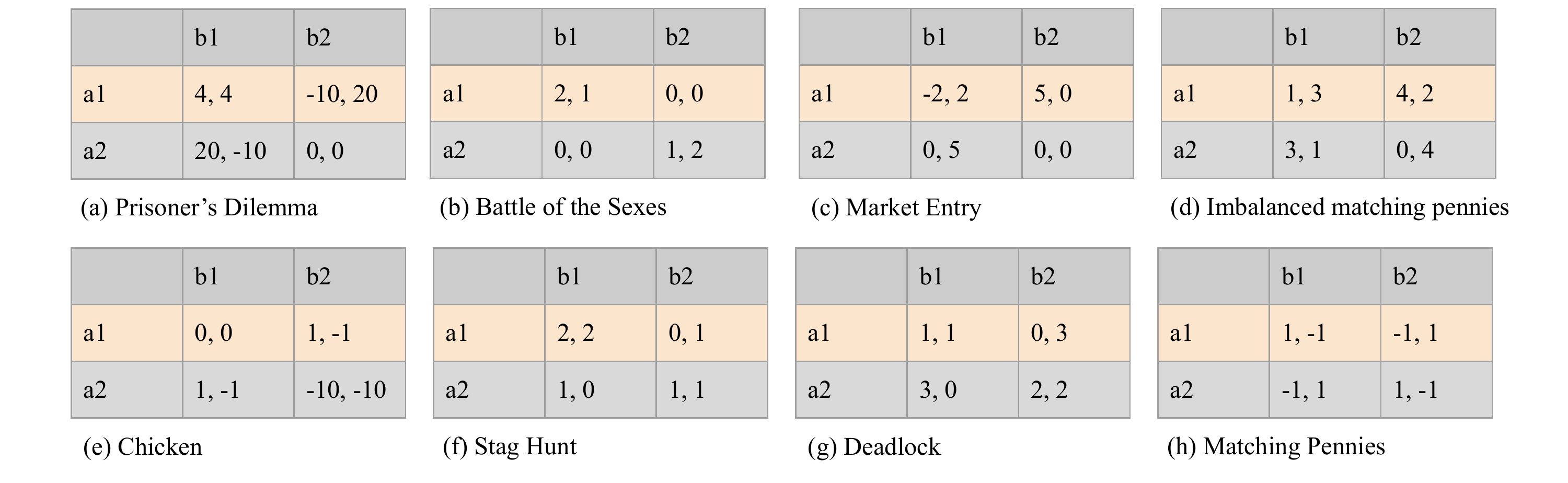}
    \caption{Various $2 \times 2$ matrix games used in our evaluations.}
    \label{afig:matrix}
\end{figure*}

For tasks with $2\times 2$ matrix games, we use the matrix games shown in \autoref{afig:matrix}. The games used in our experiments are:
\begin{itemize}
    \item Prisoner’s Dilemma: A game where both players have a dominant strategy of betraying the other, but they would both be better off cooperating. It demonstrates why cooperation can be difficult to achieve.
    \item Battle of the Sexes: A game where players have different preferences over outcomes, modeling a coordination problem.
    \item Market Entry: A game where one player must decide whether or not to enter a market based on the likely action of a competitor. It models predictive reasoning about other agents.
    \item Matching Pennies: A game where players want to match/mismatch a coin flip. It requires recognizing and exploiting the other player's incentives.
    \item Imbalanced Matching Pennies: A variant of Matching Pennies with unequal rewards, adding complexity to strategy selection.
    \item Deadlock: A game with no dominant strategies where any combination of actions leads to a tie. It demonstrates the limits of naive payoff-maximizing reasoning.
    \item Stag Hunt: A game demonstrating the tension between safety and social cooperation. 
    \item Chicken: A game modeling escalation where players must decide whether to avoid or engage in a dangerous confrontation. The first to swerve away is the "chicken".
\end{itemize}

\begin{figure*}
    \centering
    \includegraphics[width=\textwidth]{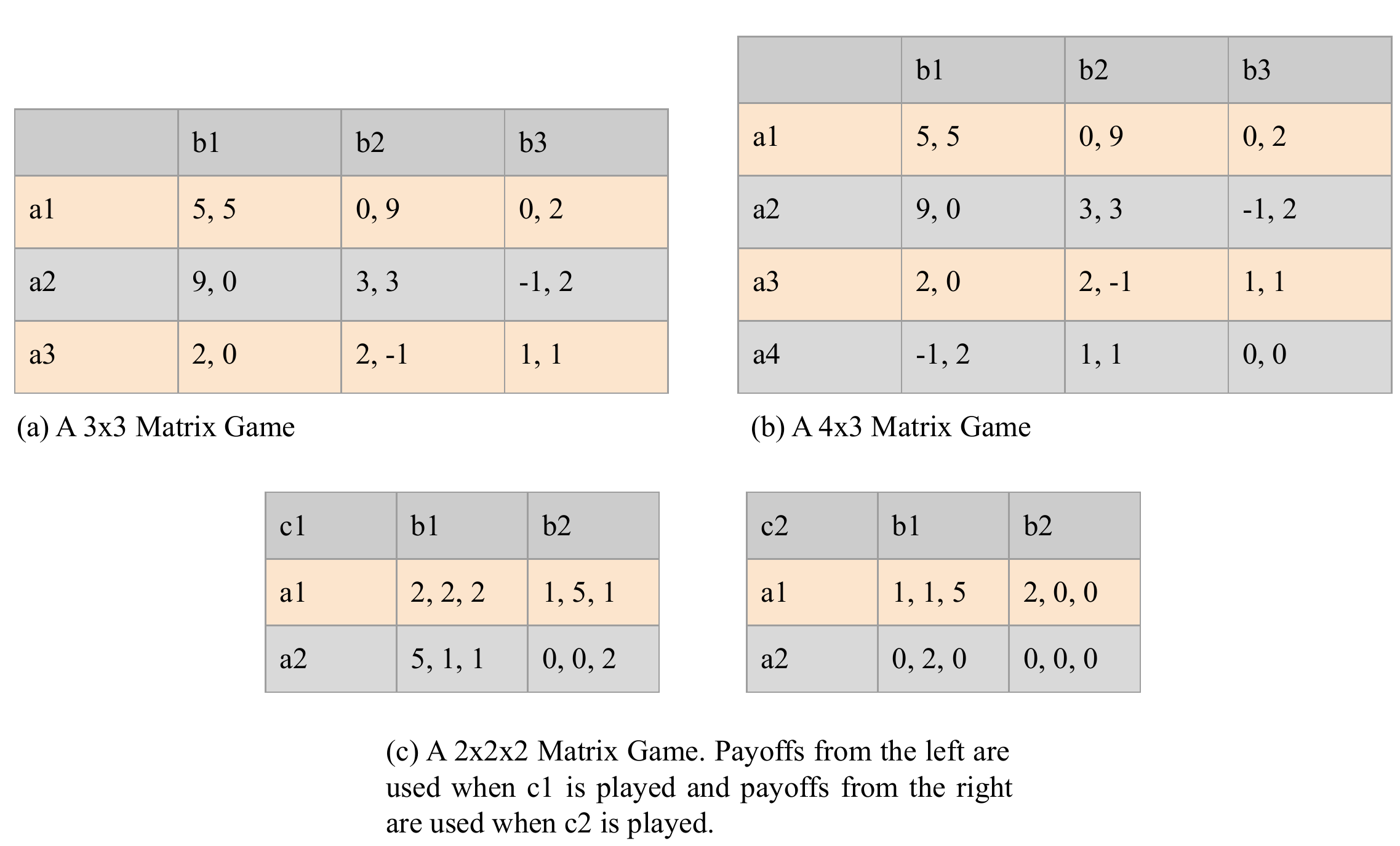}
    \caption{Matrix games with novel structures, with more actions in (a),(b) and with more players in (c).}
    \label{afig:matrix-more}
\end{figure*}

\label{asec:matrix}
\subsection{Complex Games}
\begin{figure*}
    \centering
    \includegraphics[width=0.7\textwidth]{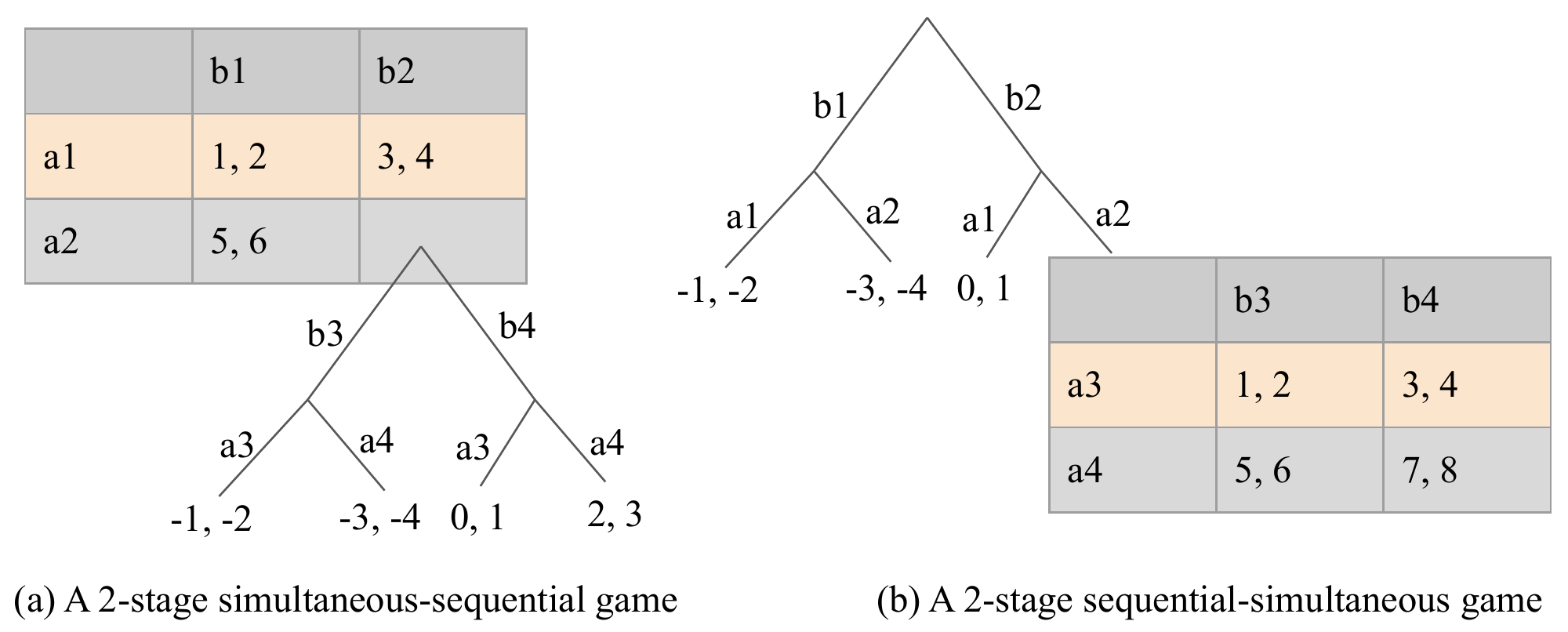}
    \caption{A visualization of a 2-stage game with different rules of playing actions in the rounds. In (a), players first play a simultaneous game and then play a sequential game if the both choose (a2, b2). In (b), players first play a sequential game and then play a simultaneous game if they both choose (a2, b2). Only games with trivial payoffs are shown here; these are used in the incontext demonstrations.}
    \label{afig:complex}
\end{figure*}

Complex 2-stage matrix games that have a sequential and simultaneous component are shown in \autoref{afig:complex}.
\label{asec:complex}

\section{Instructions given to participants}
Here are the instructions that participants were shown for the negotiation experiment.

\textbf{Description} 

You are invited to participate in a research study on strategic reasoning and negotiation with AI agents. We are interested in understanding how AI agents interact and negotiate with humans. The study will focus on how individuals negotiate with a bot and how the bot responds to different negotiation strategies. The study aims to understand how AI agents can be designed to be more effective in negotiation scenarios, by understanding the strategies that humans use and how the bot responds to them. Based on this information, we aim to develop AI agents that can adapt their behavior and improve their negotiation strategies.

\textbf{Purpose}

The purpose of this research study is to test and better understand interactions between humans and intelligent autonomous systems. 

\textbf{Procedures}

In the first phase of the experiment, you will be asked to play a negotiation game with the ai agent. Next, we will ask you for feedback about the agent’s behavior: e.g., did the agent act in a way that made sense, etc. The researchers will collect these responses, as well as data from the agent’s performance. 
If you agree to be in this study, you will be asked to do the following:

You will play negotiation games with 3 different ai agents. You will play 3 games against each of them. In each game, you have to split a pot of hats, books and balls with the ai agent. The number of these objects and their values to you and the ai agent varies across the games. On every turn, you can take 3 actions: \texttt{propose}, where you say how many objects you want to keep, \texttt{accept}, when you want to accept a proposed deal, and \texttt{reject}, when you want to reject a deal and end the game. The ai agent can take the same actions too. Every lasts for at most six turns after which you must accept or reject the deal. Your goal is to maximize the value of the items that you get. 

After this, you are asked to compare the behaviors with each other in terms of performance and capability.
At the end, you will fill in a questionnaire.
We will only record your preference responses in each run and the results of the negotiation interaction.

\textbf{Study duration}

Participation in this study will involve a total of 25 minutes of your time.

\textbf{Risks/Discomforts}

The risks and discomfort associated with participation in this study are no greater than those ordinarily encountered in daily life. 

\textbf{Confidentiality}

Your study data will be handled as confidentially as possible. If results of this study are published or presented, individual names will not be used.



\textbf{Compensation/Payment}

You will receive \$15 for your participation in the study in the form of an Amazon gift card.

\textbf{Costs}

You will not be charged for any of the study activities.

\textbf{Rights}

Participation is voluntary.
Refusal to participate will not involve penalty or loss of benefits to which subjects are otherwise entitled.
Subjects can discontinue participation at any time will not involve penalty or loss of benefits to which subjects are otherwise entitled.
You may refuse to answer particular questions.

\end{document}